\documentclass[a4paper,fleqn]{cas-dc}

\usepackage[numbers]{natbib}

\usepackage{epsfig}
\usepackage{graphicx}
\usepackage{amsmath}
\usepackage{amssymb}
\usepackage{subfigure}
\usepackage{caption}
\usepackage{xcolor}
\usepackage{enumitem}
\usepackage{rotating}
\usepackage{array}

\def\tsc#1{\csdef{#1}{\textsc{\lowercase{#1}}\xspace}}
\tsc{WGM}
\tsc{QE}
\tsc{EP}
\tsc{PMS}
\tsc{BEC}
\tsc{DE}

\newcolumntype{P}[1]{>{\centering\arraybackslash}p{#1}}

\newcommand{\boldheading}[1]{\vspace{2mm}\noindent\textbf{#1}}

\begin{document}
\let\WriteBookmarks\relax
\def\floatpagepagefraction{1}
\def\textpagefraction{.001}
\shorttitle{Automated rip current detection}
\shortauthors{de Silva et~al.}

\title [mode = title]{Automated Rip Current Detection with Region based Convolutional Neural Networks}                  



\author[1]{Akila de\texorpdfstring{\:}{Lg}Silva}[type=editor,auid=000,bioid=1]


\address[1]{University of California, Santa Cruz, CA, United States}

\author[1]{Issei Mori}

\author[2]{Gregory Dusek}


\address[2]{NOAA National Ocean Service, Silver Spring, MD, United States}

\author[1]{James Davis}

\author[1]{Alex Pang}



\begin{abstract}
This paper presents a machine learning approach for the automatic identification of rip currents with breaking waves.  
Rip currents are dangerous fast moving currents of water that result in many deaths by sweeping people
out to sea.  Most people do not know how to recognize rip currents
in order to avoid them. Furthermore, efforts to forecast rip currents are hindered by lack of observations to help train and validate hazard models. The presence of web cams and smart phones have made video and still imagery of the coast ubiquitous and provide a potential source of rip current observations.
These same devices could aid public awareness of the presence of rip currents. 
What is lacking is a method to detect the presence or absence of rip currents from coastal imagery.
This paper provides expert labeled training and test data sets for rip currents. We use Faster-RCNN and a custom temporal aggregation stage to make detections from still images or videos with higher measured accuracy than both humans and other methods of rip current detection previously reported in the literature.



\end{abstract}




\maketitle

\section{Introduction}

Rip currents are the most significant safety risk to swimmers along the coastlines of oceans, seas, and large lakes. \cite{Brighton13,brewster2019estimations,castelle2016rip}. The majority of beach goers do not know how to identify rip currents, and there is no  robust and reliable location-independent method to identify  them. Globally          
there are thousands of drownings each year due to rip currents \cite{Klein03, Lushine91}. A 20 year study by the US Lifesaving Association reports that 81.9\% of the 37,000 beach rescues each year are due to rip currents \cite{brewster2019estimations}. There has been no decline in the number of associated drowning fatalities, despite warning signs and educational material.

Rip currents are a well-studied ocean phenomenon \cite{bowen1969rip, RN118,RN170}. They are defined as strong and narrow channels of fast-moving water that flow towards the sea from beaches.
When waves break, they form a ``setup'' or an increase in mean water level.  This setup can vary along a shoreline depending on the amount or height of breaking waves.  Rip currents form as water tends to flow alongshore from regions of high setup (larger waves) to regions of lower setup
(smaller waves) where currents converge to form a seaward flowing rip.
The speed of seaward rips can be quite strong reaching 2 m/s, faster than an Olympic swimmer.
There are multiple factors that determine the location and strength of rips, such as bathymetry, wave height and direction, tide, and beach shape. Rip currents may either be transient or persistent in space and time.
Rips that are frequently found at the same location are usually indicative
of a fairly stable bathymetric feature such as a sand bar or reef, or a hard structure such as rocky outcrop, jetty or pier. These bathymetric features results in variations in wave breaking and setup leading to channelized rip current flow.
Transient or flash rips are independent of bathymetry and may move up or down the beach, and may appear or disappear. 

Lifeguards are often trained to identify rip currents. \linebreak However the majority of drownings occur on beaches without trained personnel \cite{australia2009national,branche2001lifeguard}. Posted signs can provide a warning, but there is evidence that most people do not find existing signs helpful in actually identifying rip currents \cite{Brannstrom15}. 

Experts at the National Oceanic and Atmospheric Administration (NOAA) use images and video to gather statistics about rip currents \cite{dusek2019webcat}. These data are supporting the validation of a rip current forecast model to alert people to potential hazards \cite{RN136}. The most commonly used method to visualize rip currents from video is time averaging, summarizing a video as a single image \cite{holman2007history}. In \cite{maryan2019machine} boosted cascade of simple Haar like features, a machine learning technique, was used to detect rip currents in time averaged images.  However these time averages when manually assessed can be misinterpreted. Furthermore, they are not readily available nor interpretable by the average beachgoers, and the process of averaging removes available information. 

In recent years the coastal engineering community has successfully used deep neural networks to solve many problems. Classification problems such as classifying wave \linebreak breaking in infrared imagery \cite{buscombe_data-driven_2019}, beach scene and other landscape classification \cite{buscombe_landscape_2018}, automated plankton image classification \cite{luo_automated_2018} \emph{and} ocean front recognition \cite{lima_learning_2017} were formulated as deep learning problems using convolutional neural networks.  Furthermore, some regression problems such as optical wave gauging \cite{buscombe_optical_2020}, tracking remotely sensed waves \cite{stringari_novel_2019}, typhoon forecasting \cite{jiang_deep_2018} were also solved using deep neural networks. In addition, generative adversarial networks, a type of deep neural networks, were used to improve the quality of downscaling of ocean remote sensing data \cite{ducournau_deep_2016}. 

\begin{figure*}
    \begin{center}
        \centering
           \includegraphics[width=0.32\textwidth, height =1.1in]{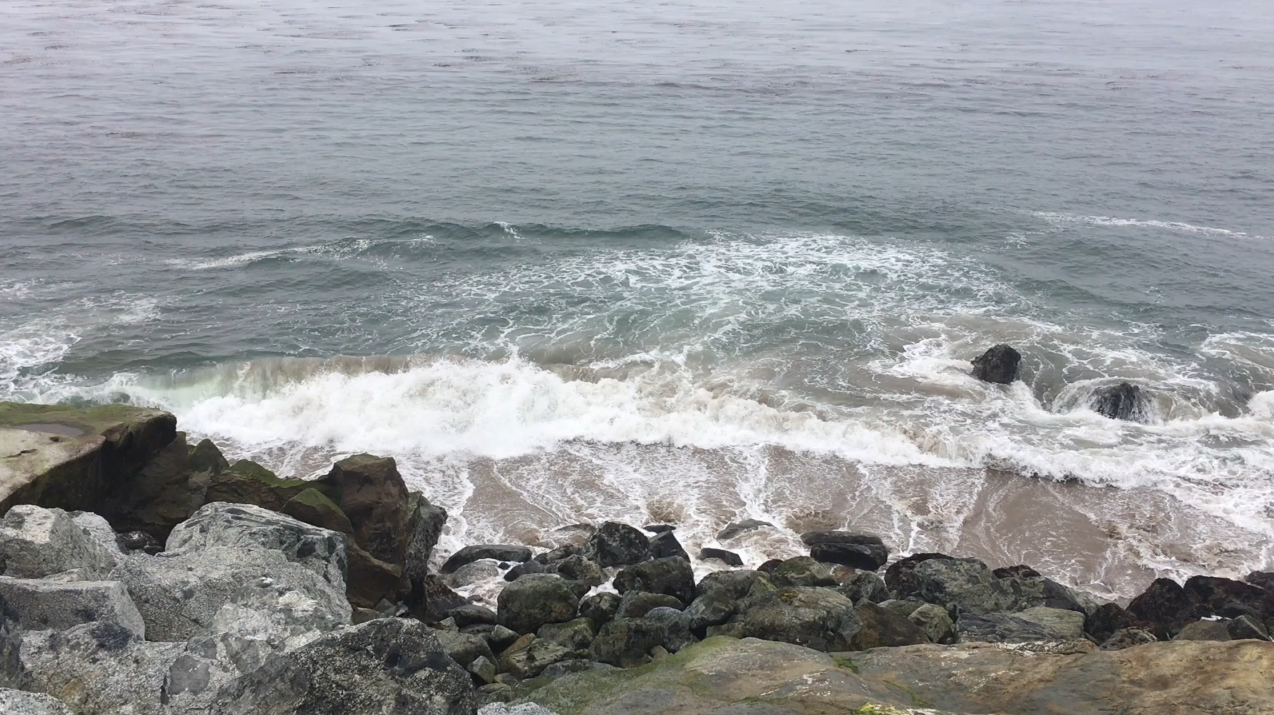}
           \includegraphics[width=0.32\textwidth, height =1.1in]{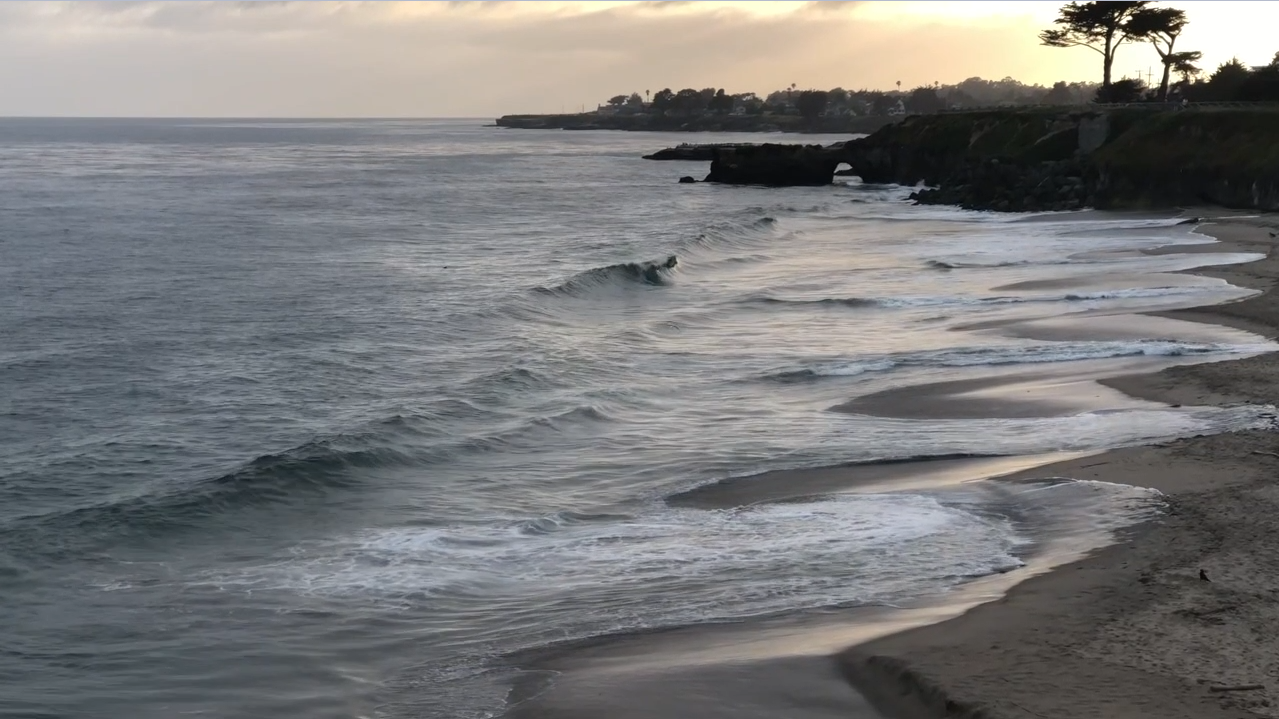}
            \includegraphics[width=0.32\textwidth, height =1.1in]{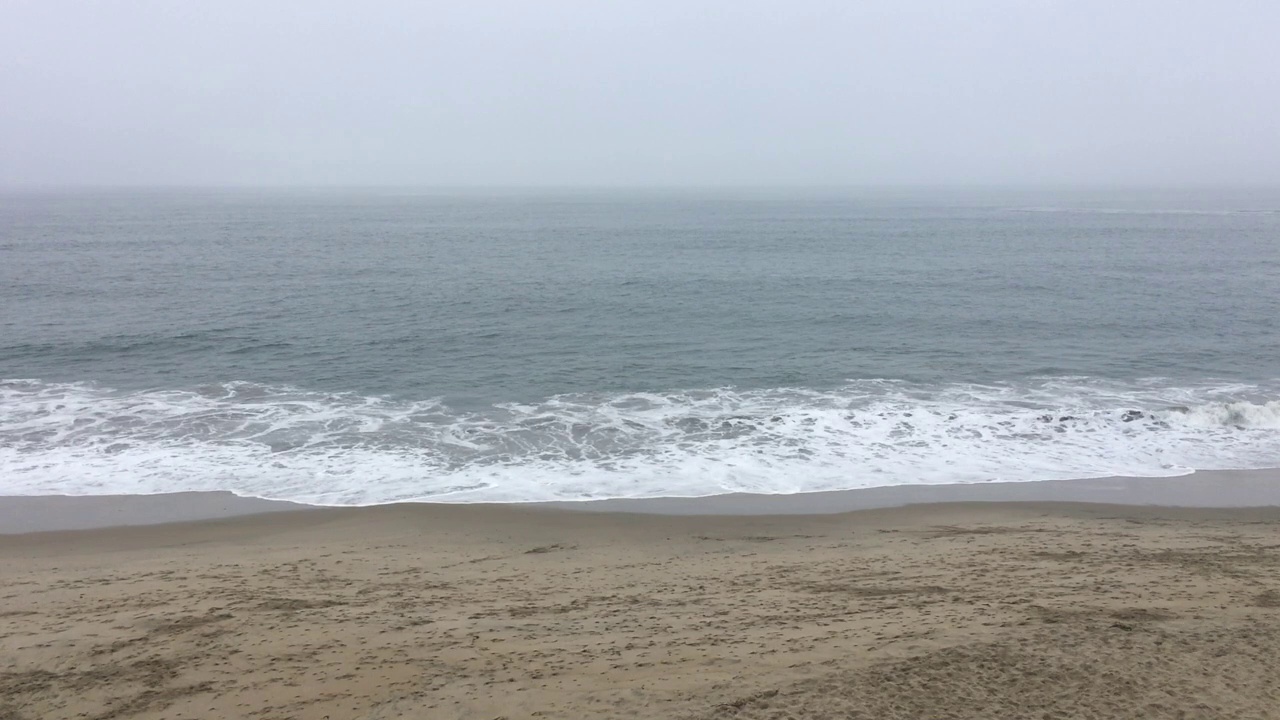}
    
    \vspace{0.5mm}
    
            \includegraphics[width=0.32\textwidth, height =1.1in]{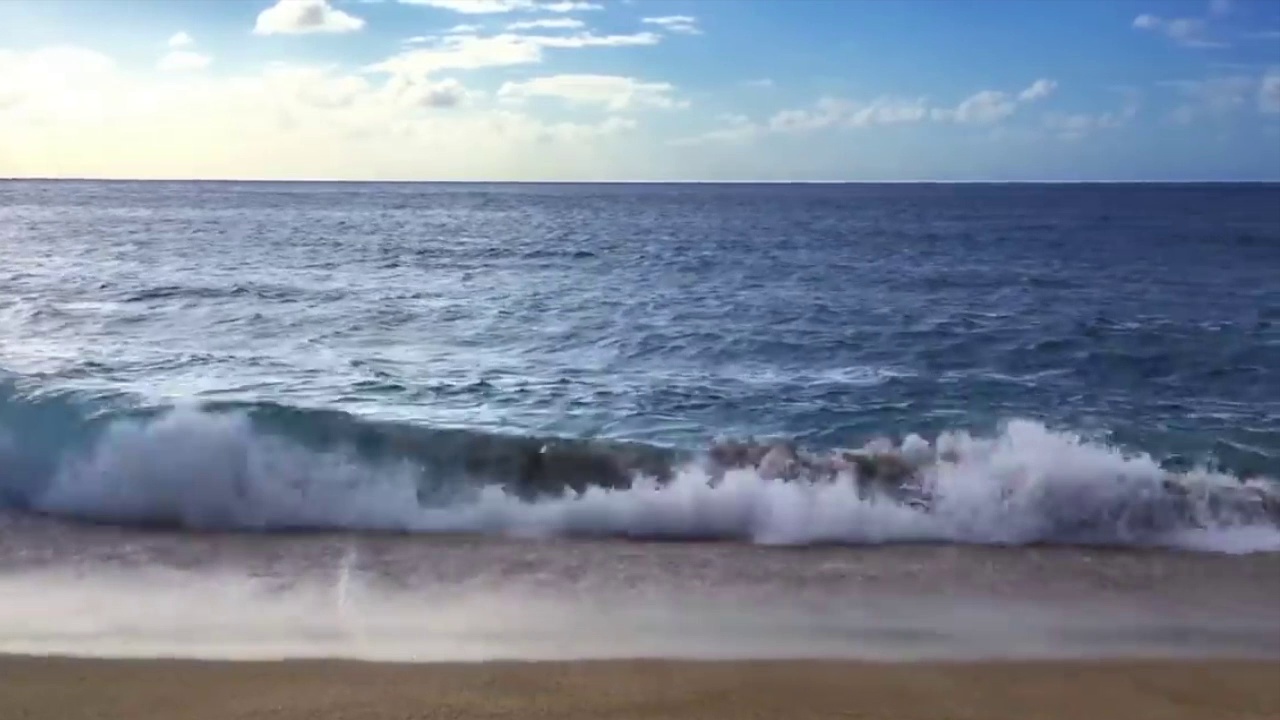}
            \includegraphics[width=0.32\textwidth, height =1.1in]{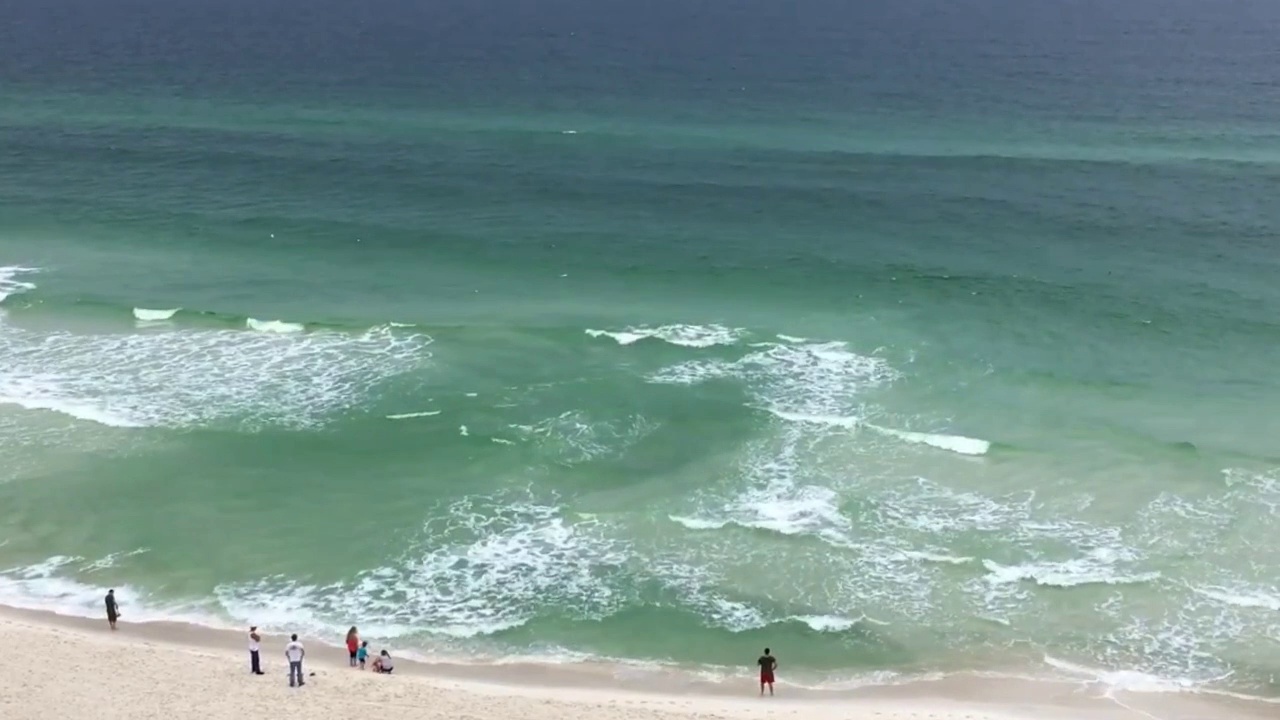}
            \includegraphics[width=0.32\textwidth, height =1.1in]{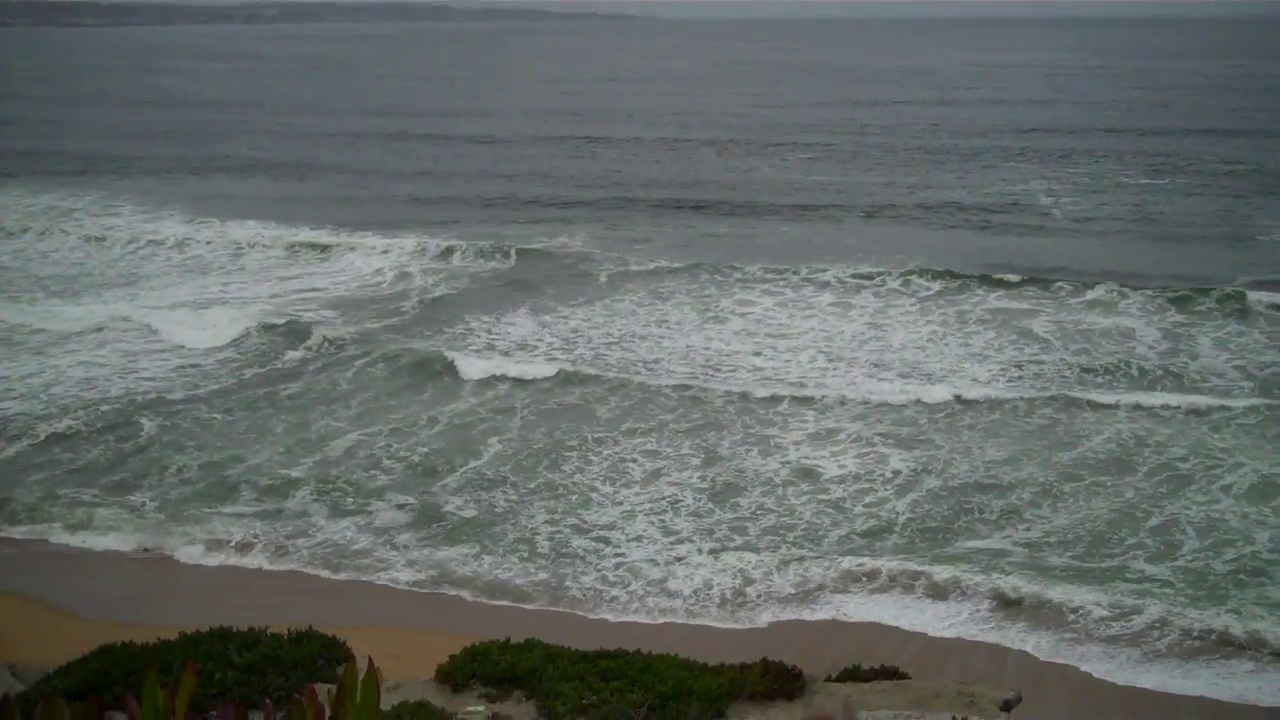}
             \captionof{figure}{A collection of beach scenes, some of which contain dangerous rip currents. Unfortunately these ``objects'' do not have clear shape, and most people find them hard to identify. This paper describes a computer vision system with detection accuracy higher than both existing published methods and human observers.} \label{teaserfig}
    \end{center}
\end{figure*}

Object detection with deep neural networks is well  \linebreak studied in the computer vision community. However most benchmarks and research focus on detecting physical \linebreak objects with boundaries between what is and is not \linebreak an object \cite{deng2009imagenet,everingham2010pascal,lin2014microsoft}. Rip currents are ephemeral ``objects'' which are not observable in every frame, and amorphous without clearly defined boundaries even when observable. It is not obvious whether existing methods are applicable. \linebreak Figure~\ref{teaserfig} provides a set of examples, illustrating the difficulty of the problem.

Our work is aimed at introducing this problem to the
coastal engineering community, and showing that object detection methods \emph{are} applicable. We gathered training data sets of rip currents and worked with experts at NOAA to ensure that test data were labeled correctly. We use Faster-RCNN~\cite{ren2015faster} with a custom temporal aggregation stage that allowed us to achieve detection accuracy that is higher than both humans and other methods of rip current detection previously reported in the literature.

\vspace{1mm}
\noindent
The contributions of this paper are: 
\begin{itemize}
\itemsep0em
    \item Evidence that region based convolutional neural networks (CNN) approach for object detection is applicable to amorphous and ephemeral objects such as rip currents
    \item Analysis showing rip current detection accuracy above existing \linebreak published methods 
   \item Data sets of rip current images and video for training and testing
\end{itemize}

The remainder of the paper is organized as follows. All the related work is summarized in Section \ref{s:related_work}. We discuss how the data was collected in Section \ref{s:data_sets}. Our method is discussed in \ref{s:methods}. Results are discussed in \ref{s:results}. Limitations and discussion are in Section \ref{s:limitations_and_discussion}. In Section \ref{s:conclusion} we conclude our paper. And in Appendix \ref{s:appendix} we provide the link to the supplementary materials.

\section{\label{s:related_work} Related Work}

Rip currents are most commonly studied using in situ techniques or instrumentation. Fluorescein dye is commonly released into the ocean and the dispersion observed \cite{brander2014dye, Clark10,clark2014aerial,Pritchard60}.  Wave sensors, acoustic velocimeters, or current profilers can  be deployed at specific locations \cite{Elgar01,Inch14,Johnson04,liu2019lifeguarding}. Floating drifters with embedded GPS units have also been used to measure currents~\cite{Castelle14,castelle2016rip,schmidt2003gps}.
These methods are costly, time consuming, require technical expertise and are generally only applicable to highly localized instances in time and space. These limitations severely hinder the applicability of such approaches to both public warnings and model validation.

Time averaged images are a routine method for analyzing video in oceanic research, with 10 minutes being a common integration period~\cite{holland1997practical,holman2007history,lippmann1989quantification,pitman16}. This method is popular because averages often make identification of rip channels easier for the human eye. 
While these images are usually intended for human interpretation, Maryan et. al. apply machine learning to recognize rip channels in time averaged images \cite{maryan2019machine}.  Nelko also used time averaged images and noted that prediction schemes developed at one beach location may not be directly applicable to another without some modifications~\cite{nelko14}. In contrast to these works,
the analysis in this paper suggests that object detection on individual frames outperforms time averaged images.

Dense optical flow \cite{Barron1994,Horn1981} has been used to detect rip currents in video \cite{eurovisshort.20161155}. This method is attractive since optical flow fields can be directly compared against ground truth flow fields obtained from in situ measurements \cite{derian2017wavelet}. Unfortunately these methods are sensitive to camera perturbation, and have difficulty in areas lacking textural information. The results in this paper suggests that object detection on individual frames outperforms previous optical flow based methods.

Certain kinds of rip currents are characterized by visible sediment plumes. These can be segmented based on changes in coloration. For example, Liu et al, use thresholding in HSV color space to detect rip currents \cite{liu2019lifeguarding}.
Unfortunately, not all rip currents contain sediment plumes, and thus this method is not applicable to our data sets.

Object detection in images is well studied in the computer vision literature
\cite{han2018advanced,liu2018deep,Papageorgiou98,Redmon16,Viola2004}. These methods have been extended to detect objects in videos \cite{Han16,Kang16,zhu2017flow}. This work has not previously been applied to rip currents both because it is not clear there is an ``object'' to detect, and because of a lack of publicly available data sets for training and testing. 

This paper contributes labeled data sets for rip current detection, and evidence that object detection outperforms existing published methods on this application.

\begin{figure*}
        \centering
        \begin{turn}{90} \hspace{6.2mm}With Rips   \end{turn}
        \includegraphics[width=0.158\textwidth, height =1.1in]{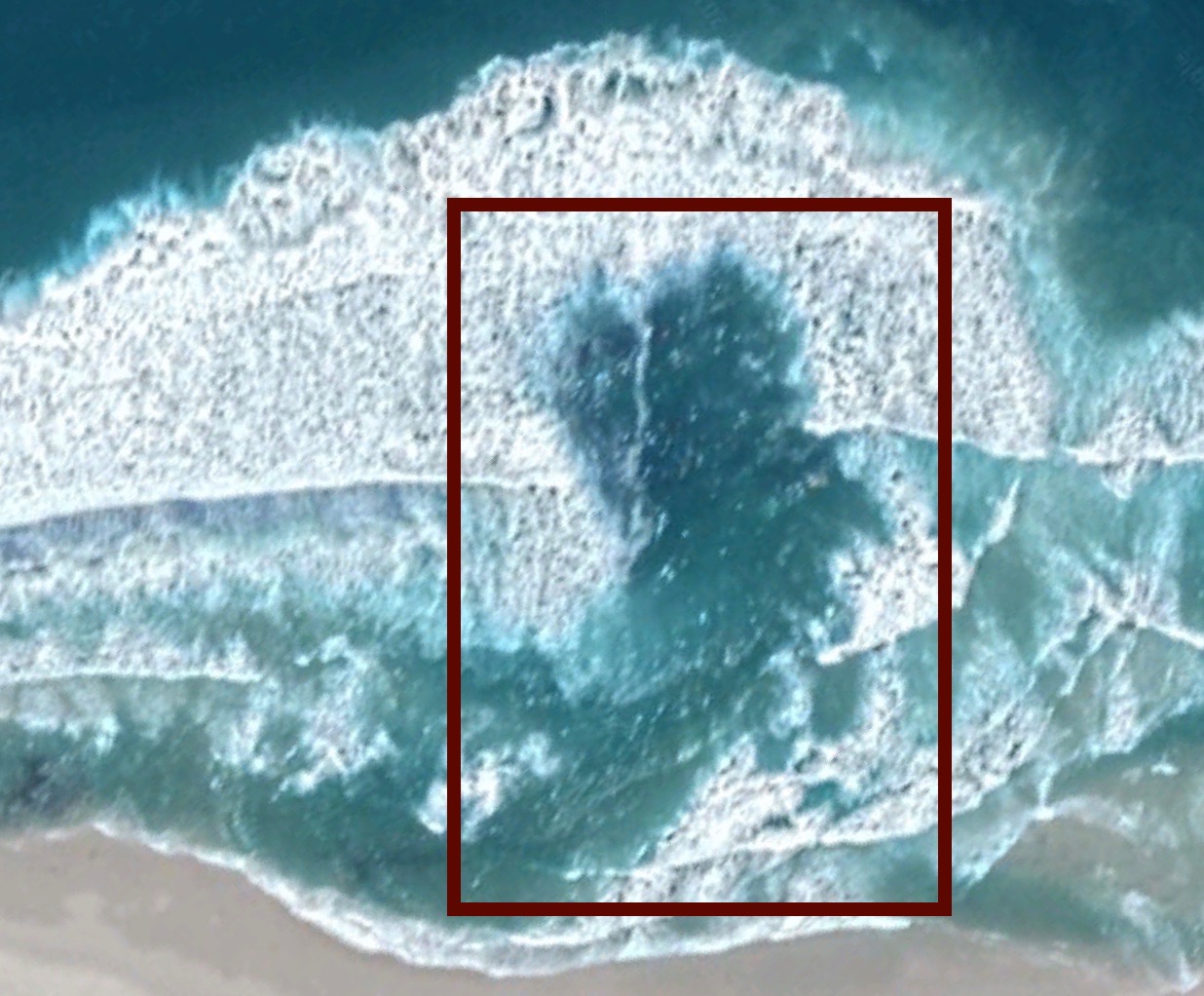}
        \includegraphics[width=0.158\textwidth, height =1.1in]{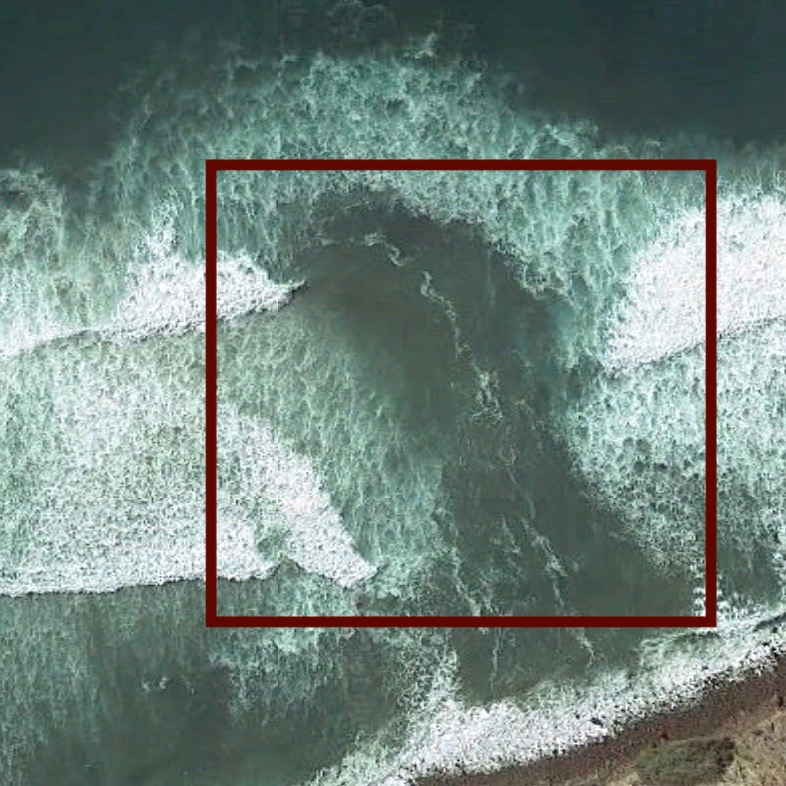}
        \includegraphics[width=0.158\textwidth, height =1.1in]{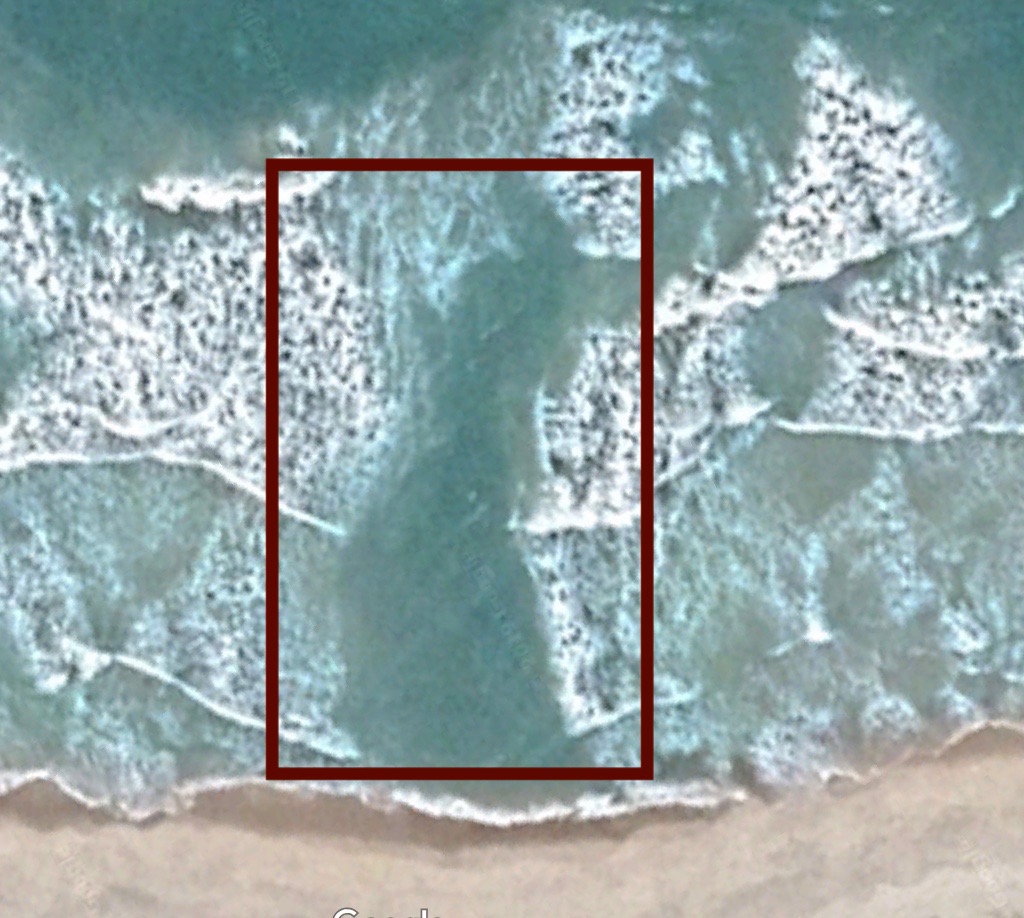}
        \includegraphics[width=0.158\textwidth, height =1.1in]{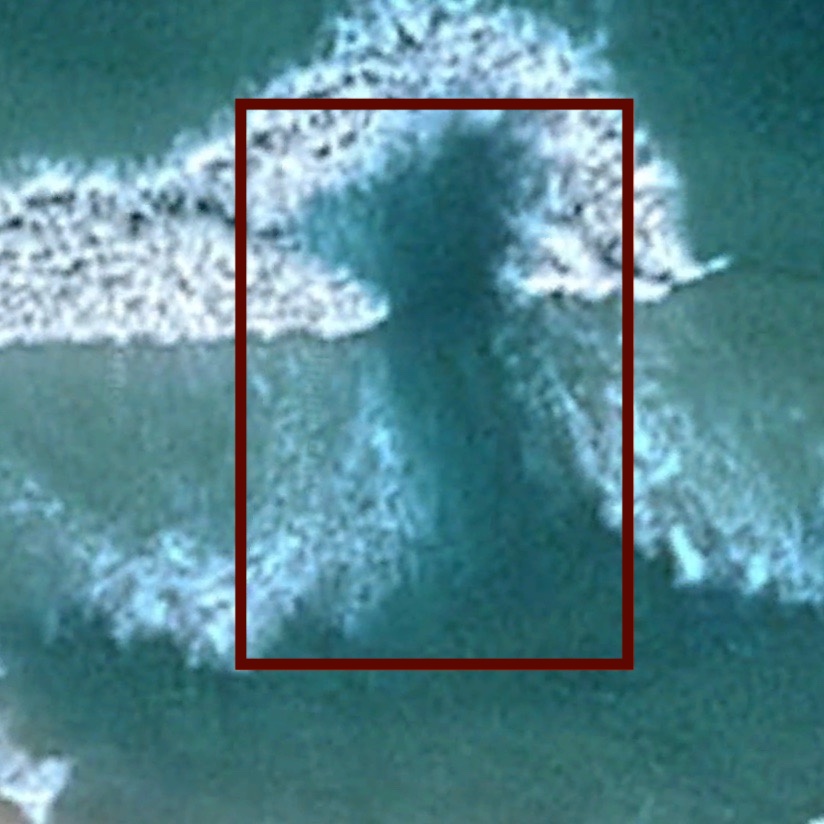}
        \includegraphics[width=0.158\textwidth, height =1.1in]{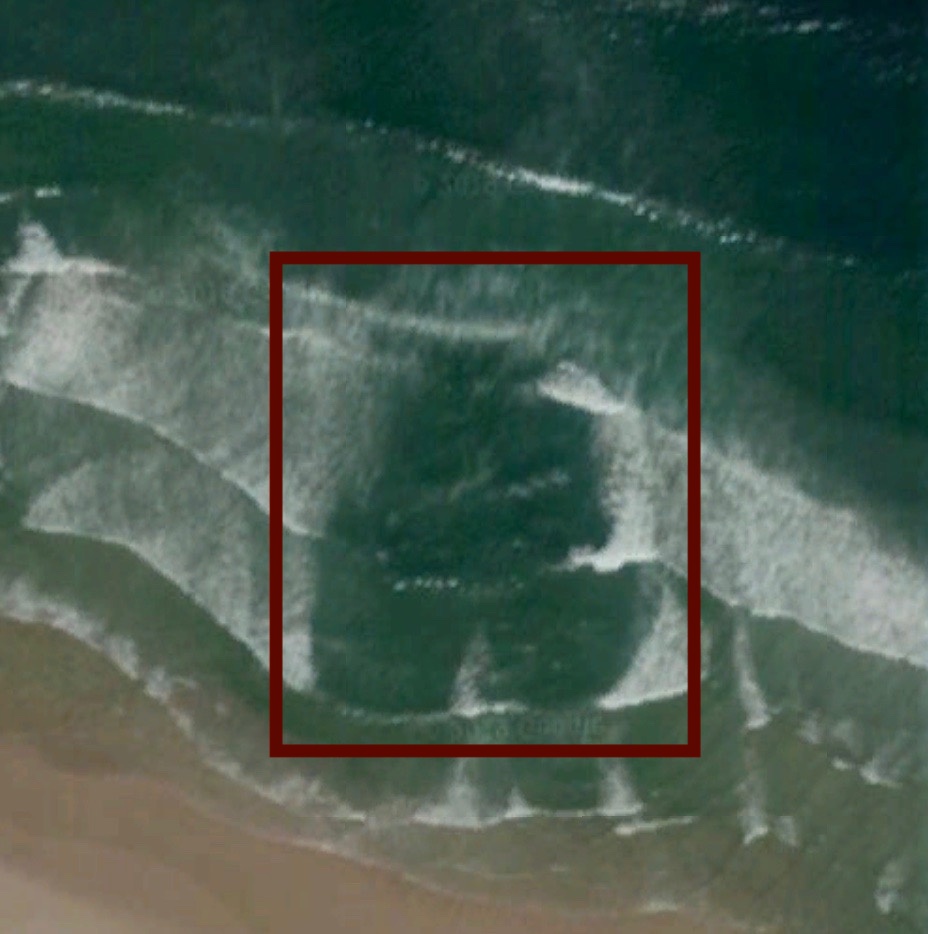}
        \includegraphics[width=0.158\textwidth, height =1.1in]{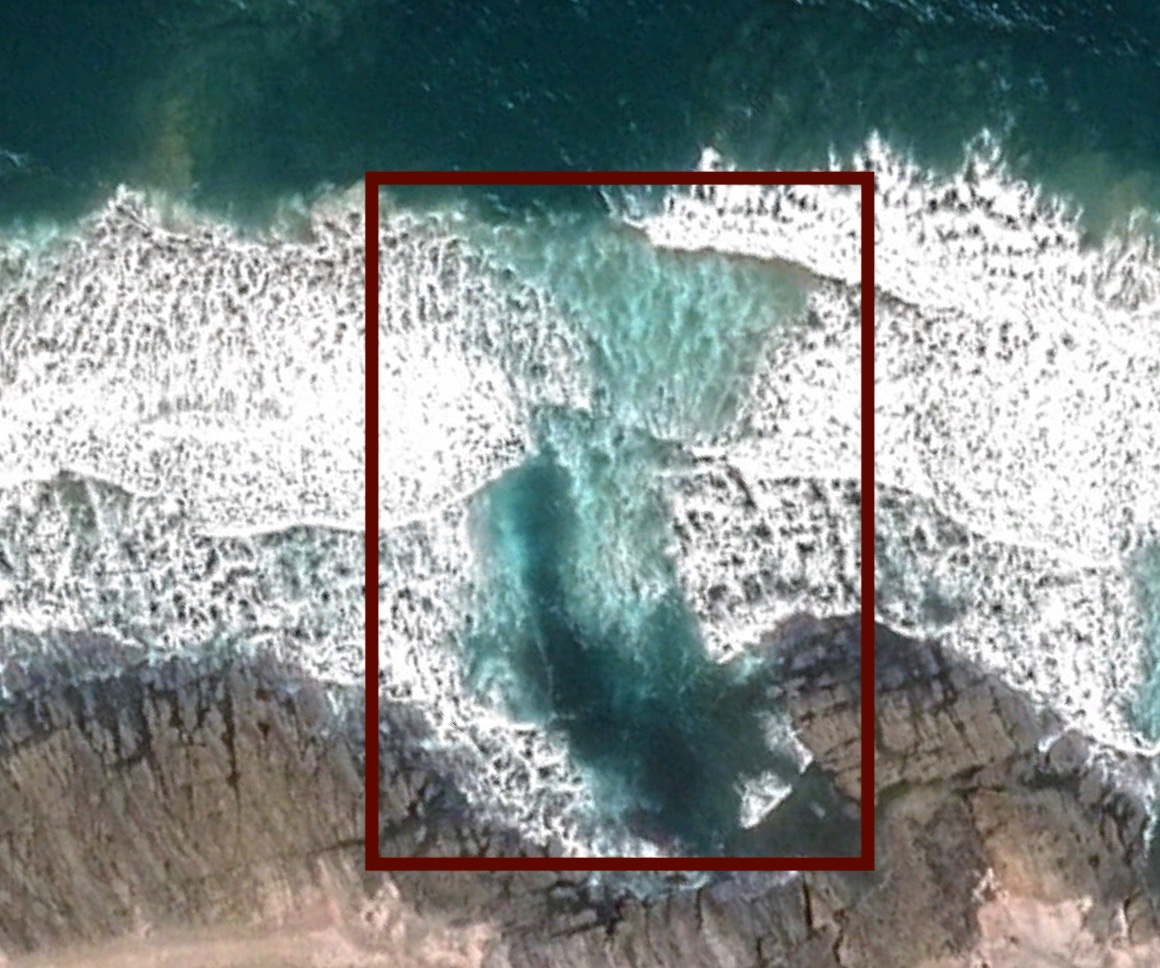}
    
        \begin{turn}{90}\hspace{5mm}Without Rips\end{turn}
        \includegraphics[width=0.158\textwidth, height =1.1in]{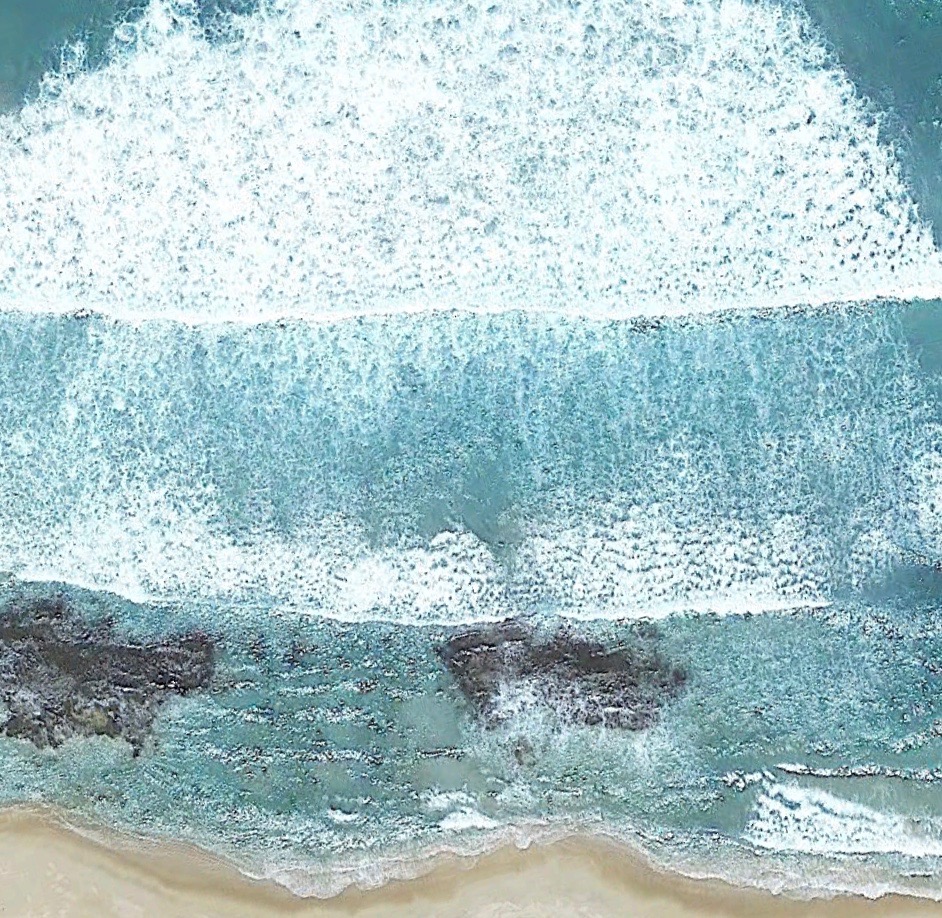}
        \includegraphics[width=0.158\textwidth, height =1.1in]{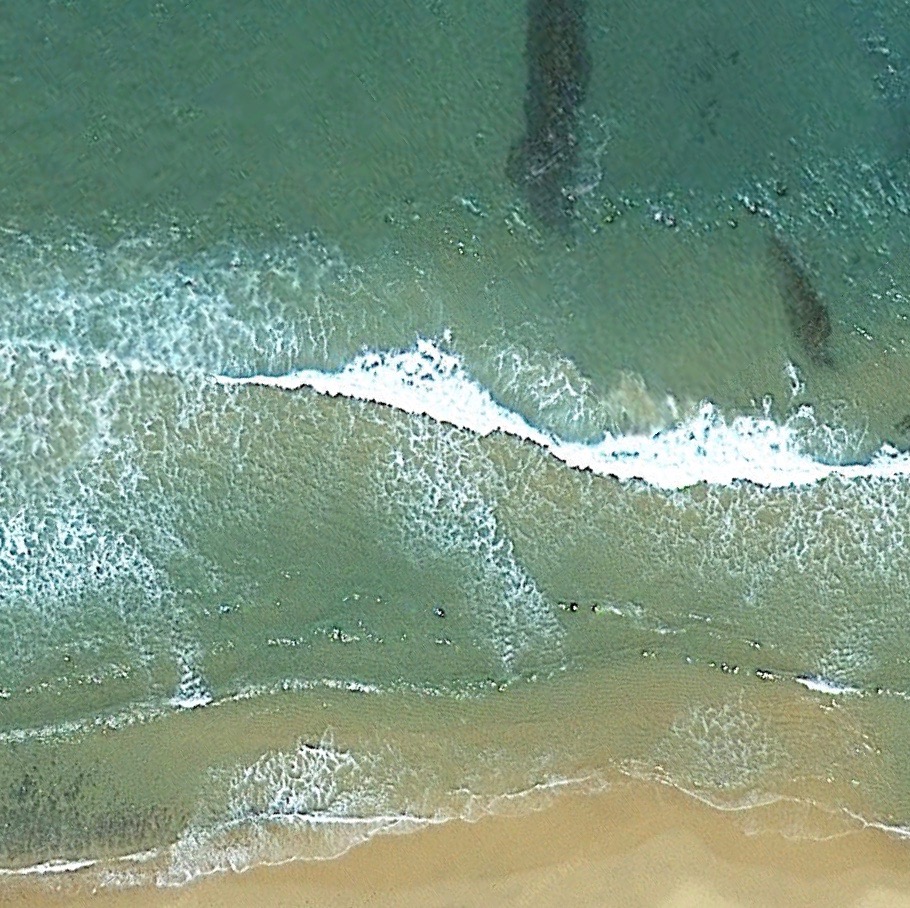}
        \includegraphics[width=0.158\textwidth, height =1.1in]{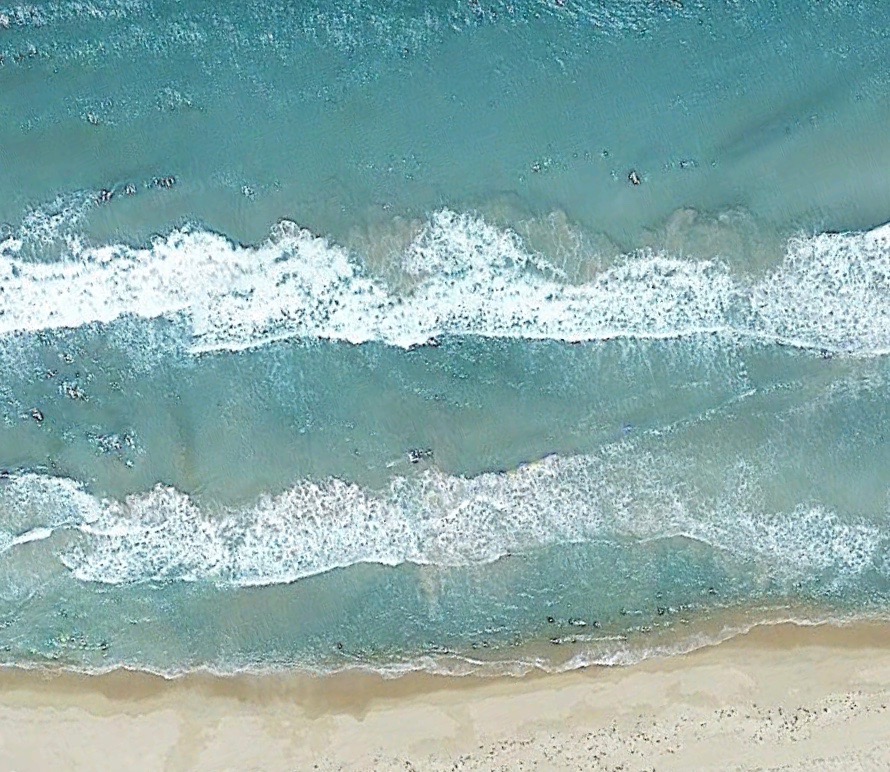}
        \includegraphics[width=0.158\textwidth, height =1.1in]{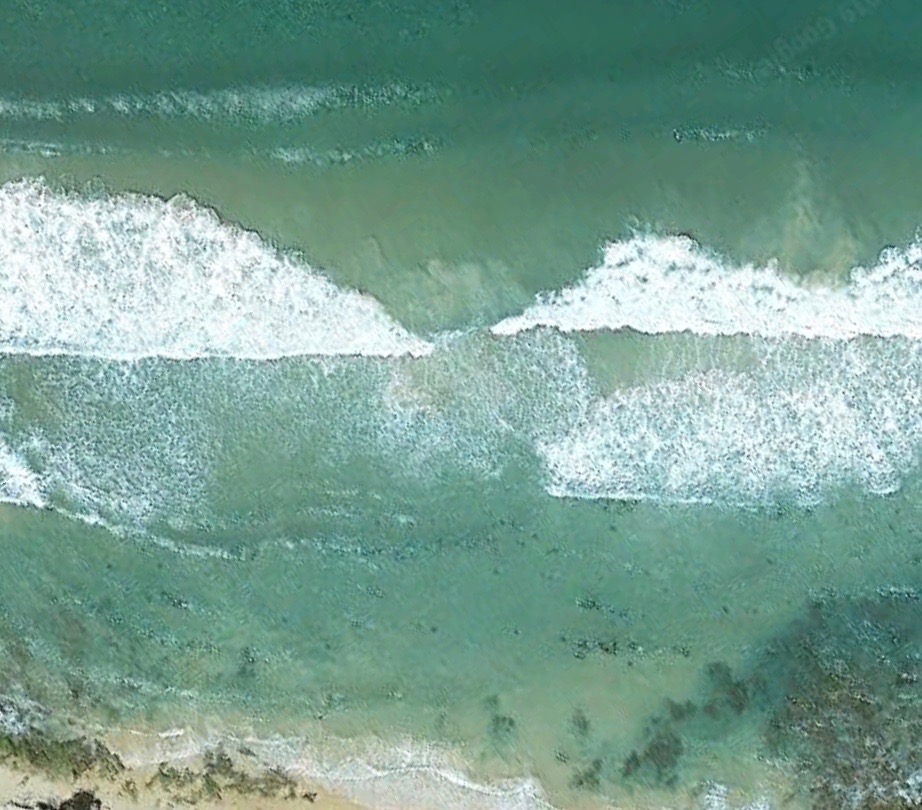}
        \includegraphics[width=0.158\textwidth, height =1.1in]{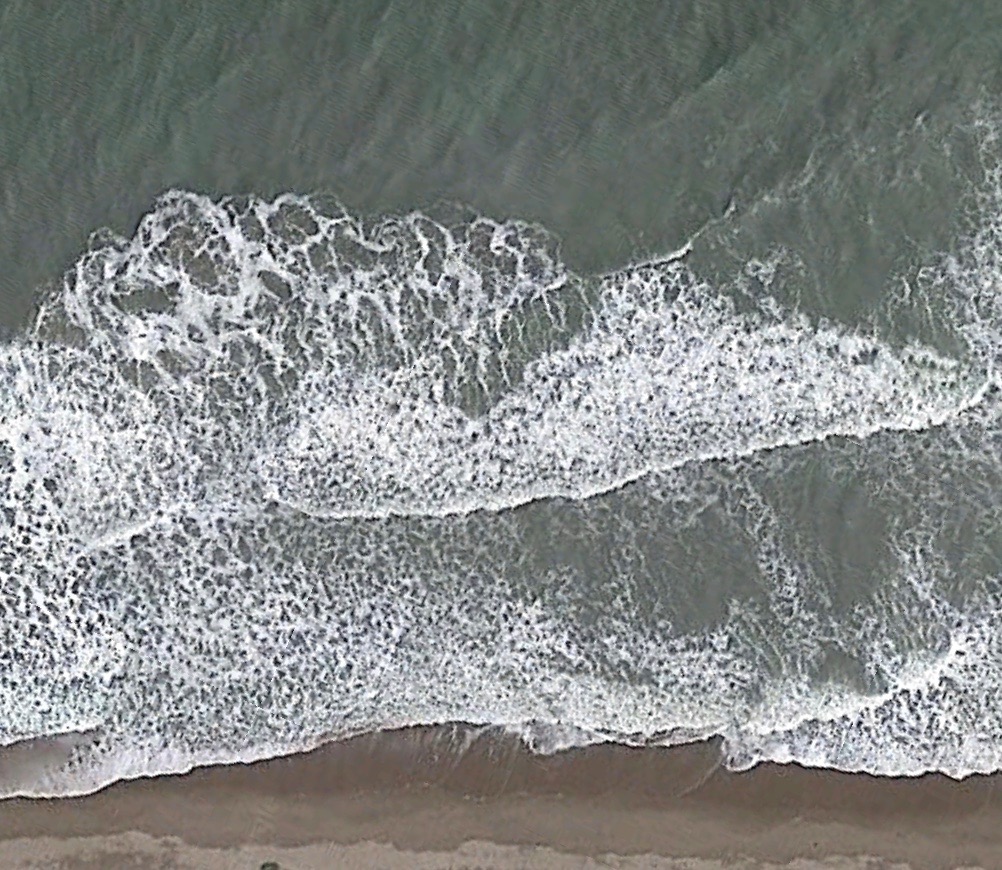}
        \includegraphics[width=0.158\textwidth, height =1.1in]{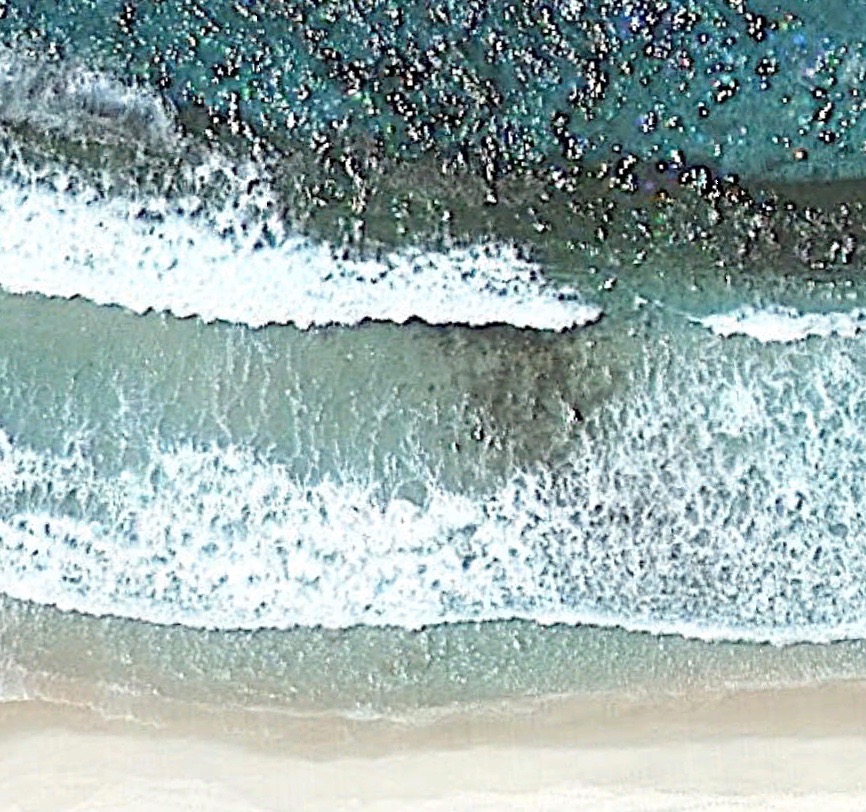}
        \caption{Examples drawn from the 2440 images we collected and labeled to build a training data set. Ground truth bounding boxes are shown in red.} 
        \label{no_rip_current_examples}
\end{figure*}  

\section{\label{s:data_sets} Data sets}

\subsection{\label{s:training} Training Data Images}

Since rip currents are a new problem domain for computer vision, we did not find any existing public databases of rip current images.  Therefore, we assembled a training data set of rip current images and non-rip current images from scratch. Our primary source for the database was Google Earth$^{TM}$, which allowed us to extract high-resolution aerial images of rip currents and non-rip currents. In total, the database contains 1740 images of rip currents and 700 images of similar beach scenes without rip currents. The images range in size from $1086 \times 916$ to $234 \times 234$ . We annotated ground truth in the rip current images with axis-aligned bounding boxes.  Some examples of the training data set are shown in Figure~\ref{no_rip_current_examples}. Note that this data set contains unambiguous easy examples. 
We used this static image data set for training models described in Section~\ref{s:methods}.

\subsection{Test Videos}

We also collected a data set of $23$ video clips consisting of $18042$ frames. There are a total of $9053$ frames with and $8989$ frames without rip currents. Image size varies from $1280 \times 720$ to $ 1080 \times 920 $ . Ground truth annotations were verified by a co-author who is also a rip current expert at NOAA. Figure~\ref{teaserfig} contains both positive and negative example frames, as well as a few frames from the training set that might be mistaken as containing a rip current. Note that these sequences contain more difficult cases. The rip currents are not visible in every frame, but a positive bounding box is applied whenever viewing the entire video segment indicates that a rip current is present.

The frames of this video data set were used for testing.
Note that the static images in the training set were taken from high elevation
while the videos used to test the model were taken from a lower perspective.
Even so, the trained model performed well on the test frames from the video collection.

In order to encourage progress in this domain, both data sets will be made available to the public. We are including a link to the data set in the supplementary materials in appendix \ref{s:appendix}. 

\section{\label{s:methods} Method}

\subsection{Static Image Detector: Faster RCNN }

  Region-based convolutional neural networks have \linebreak  achieved  great success in object detection problems.  These object detection models usually consist of separate classification and localization networks with a shared feature extraction network. We use Faster RCNN which is also composed of two components. The first is the deep convolutional neural network that proposes regions. The second is the Fast RCNN detector \cite{girshick2015fast}. 
  
  Faster RCNN follows the traditional object detection  \linebreak pipeline. It first generates region proposals, and then categorizes each proposal as either rip current or background. Secondly, the classified bounding boxes are further refined. Essentially, the model learns a mapping from the generated regions to the actual ground truth with a regression network. The model then uses this mapping ``function'' during testing to refine the generated region. These refined bounding boxes can be anywhere in a frame as features are translation invariant \cite{ren2015faster}. If there is more than one bounding box detected in a frame, we only keep track of the largest one and ignore any additional boxes.
  
  We trained the Faster RCNN model with the static image data set. Before training, each image was augmented by rotating $90^{\circ}$ degrees clockwise and counter clockwise, producing a training data set three times the size of the original training data set. All the training data was resized to $300 \times 300$ before training to save computation time. 
  


\subsection{Frame Aggregation}

\begin{figure}
\begin{center}
   \includegraphics[width=1.0\linewidth]{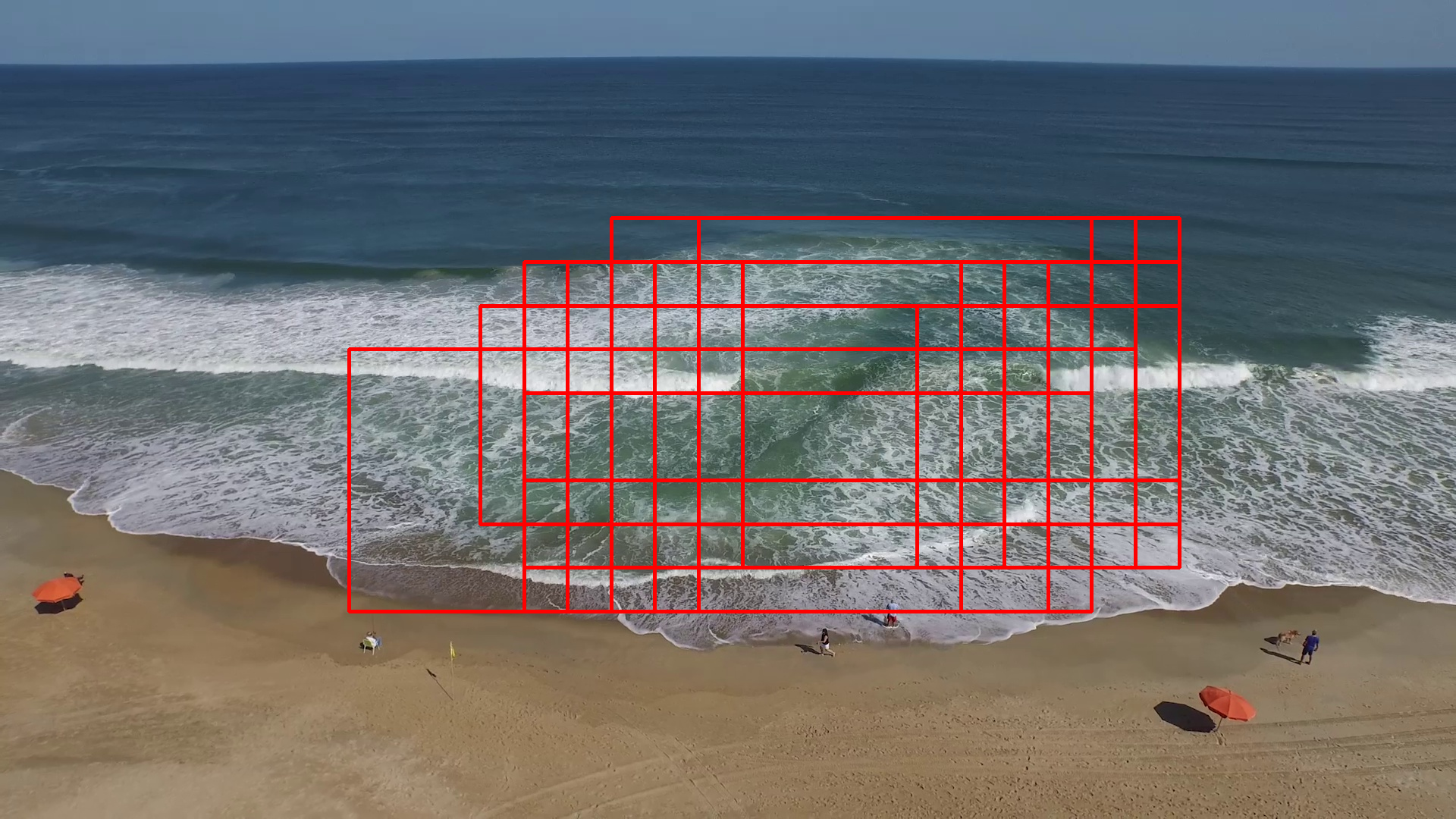}
\end{center}
   \caption{Amorphous phenomena without clear boundaries, like rip currents, result in wide variation in detected bounding boxes. All detected bounding boxes from the frames of one video are superimposed onto a single image. Bounding boxes may differ significantly even in consecutive frames.}
\label{fig:multi_detections}
\end{figure}

Static object detection models only consider the information in the frame currently being processed. 
However, rip currents are natural ocean phenomena, with shape and texture change depending on many external factors such as weather, wind speed, wave field characteristics, water flow speed, floating debris, and dirt sediments. The exact boundaries of a rip current are not well defined. This is different than objects with well-defined edges such as pedestrians or vehicles. Applying detection algorithms to objects with amorphous boundaries such as rip currents produces bounding boxes with variable sizes and locations in adjacent video frames.  In Figure \ref{fig:multi_detections} we illustrate this variability by drawing all correctly detected bounding boxes from one video sequence onto a single frame. 

This variability affects overall accuracy, and would not instill confidence in the results if these bounding boxes were presented to a user as a video overlay. Thus we investigate temporal smoothing and aggregation to improve the results. 

\begin{figure}
\begin{center}
   \includegraphics[width=1.0\linewidth]{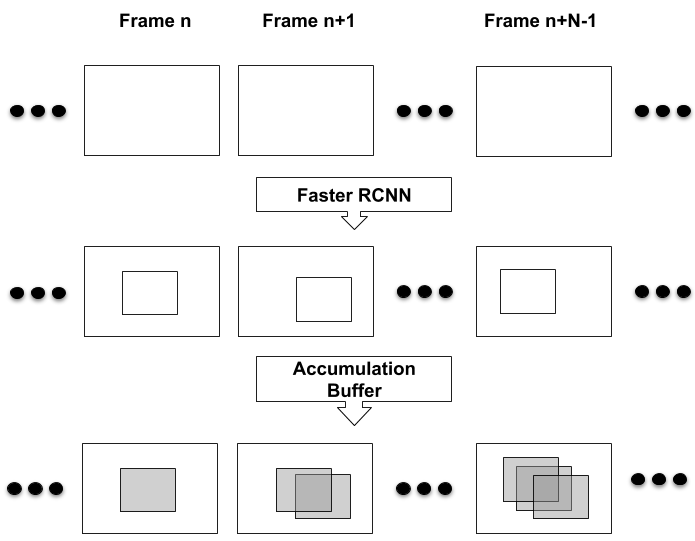}
\end{center}
  \caption{Frame aggregation for a window of length $N$. First row shows the input frame sequence. Second row shows the detections from Faster RCNN.  Third row shows the accumulation buffer. }
\label{fig:frame_aggregation}
\end{figure}

\begin{figure}
\begin{center}
  \includegraphics[width=1.0\linewidth]{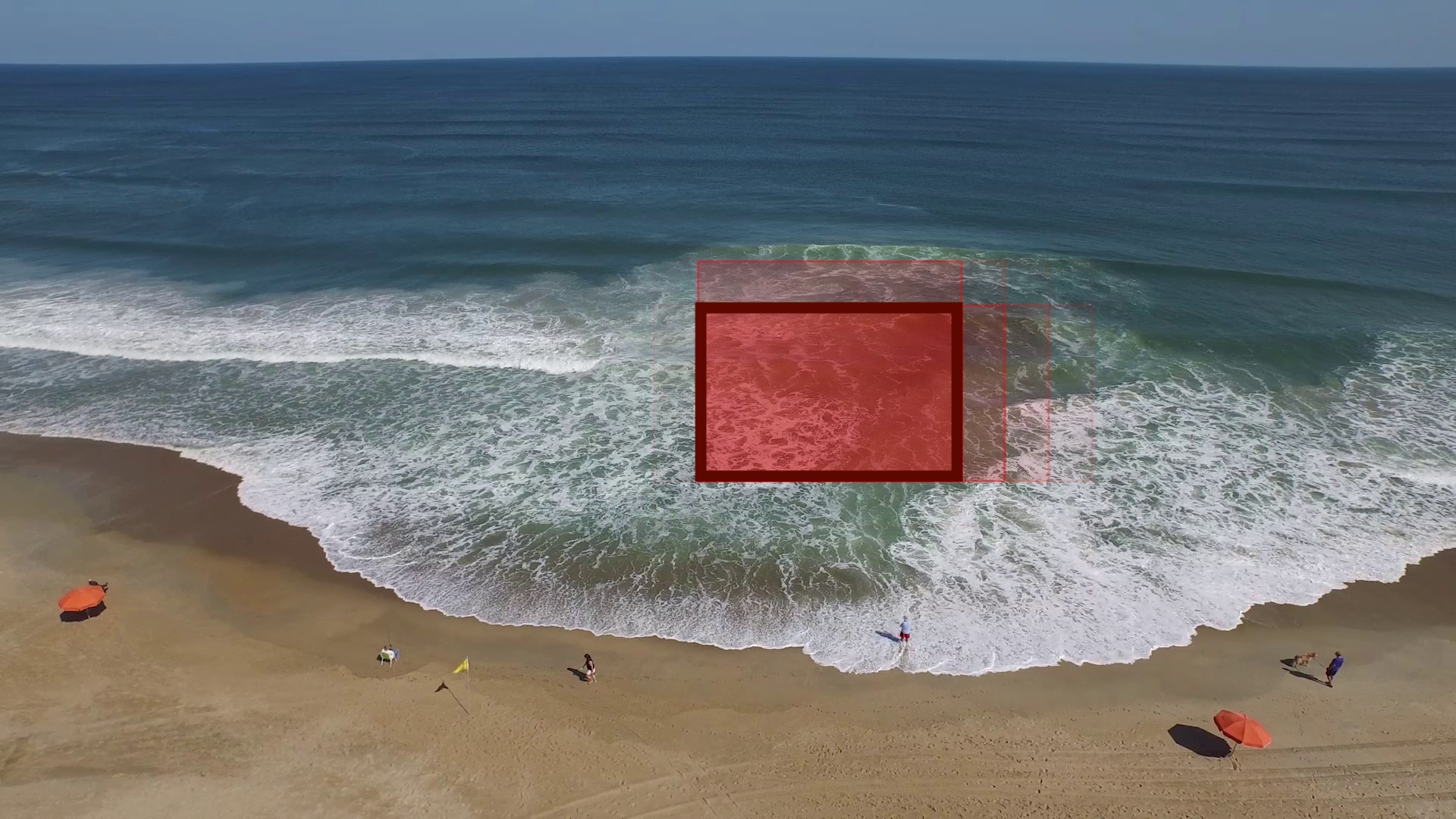}
\end{center}
  \caption{A visualization of the resulting accumulation buffer values. Regions with a higher value are shown in more opaque red. The resulting bounding box around thresholded values is shown in solid dark red. The full video can be seen in the supplementary material in appendix  \ref{s:appendix}   }
\label{fig:buffer}
\end{figure}

\begin{figure}
\begin{center}
  \includegraphics[width=1.0\linewidth]{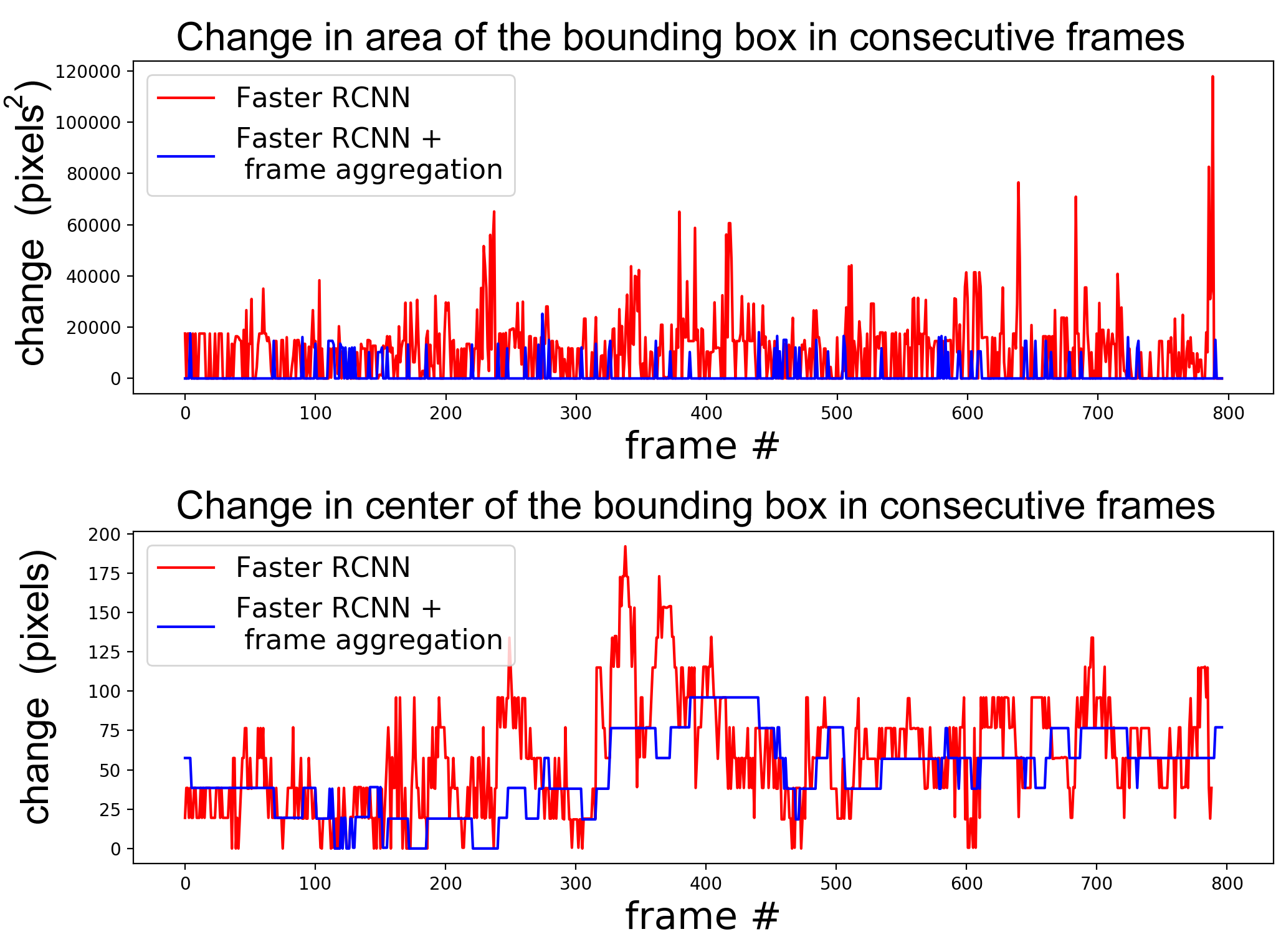}
\end{center}
  \caption{Plots showing the differences in area and center of bounding boxes in consecutive frames. Without frame aggregation (in red) there are much higher differences in bounding box sizes and positions than there is after frame aggregation (in blue).}
\label{fig:diff_bbox_size}
\end{figure}


We find the overlapping regions of the detected bounding boxes by using an accumulation buffer with the same size as the input frame and initialized as a zero matrix. We consider a temporal window of $N$ frames to build the accumulation buffer. In the first $N-1$ frames, the accumulation buffer is incremented by 1 for each region within a detection bounding box.  Starting with frame $N$, the area covered by a detection bounding box is incremented by 1 and capped at a maximum of $N$.  Regions not covered by a detection bounding box are decremented by $1$, but retains a minimum of 0 in the accumulation buffer. In effect, the accumulation buffer keeps track of the bounding boxes using a sliding window of $N$ frames. The process of building the accumulation buffer is illustrated in Figure \ref{fig:frame_aggregation}, where areas with higher values are displayed as darker regions in the accumulation buffer. 
For purposes of identifying a single bounding box over the collection of bounding boxes across $N$ frames, we consider only the parts of the accumulation buffer where the value is at least $T$, and draw the tightest possible axis-aligned bounding box around this region (see Figure~\ref{fig:buffer}). This is the aggregated detection. 
In our implementation we use N=60 and T=50.

Before frame aggregation, large variations in bounding box size occur in almost all consecutive frames, shown in Figure \ref{fig:diff_bbox_size} top. After frame aggregation, the average size \linebreak change is much smaller, with most frames having zero \linebreak change in size from the prior frame.
The variation in position of the bounding boxes is similarly reduced by frame aggregation, as seen in Figure \ref{fig:diff_bbox_size} bottom.
This improved temporal coherence provides a smoother and more consistent portrayal of the rip current extent when shown as an overlay on the video. 

\begin{figure*}
\begin{center}
    \centering
       \includegraphics[width=0.32\textwidth, height =1.1in]{images/teaser/no_rip_04.png}
       \includegraphics[width=0.32\textwidth, height =1.1in]{images/teaser/no_rip_06.png}
        \includegraphics[width=0.32\textwidth, height =1.1in]{images/teaser/no_rip_01.jpg}
        \includegraphics[width=0.32\textwidth, height =1.1in]{images/teaser/18463.jpg}
        \includegraphics[width=0.32\textwidth, height =1.1in]{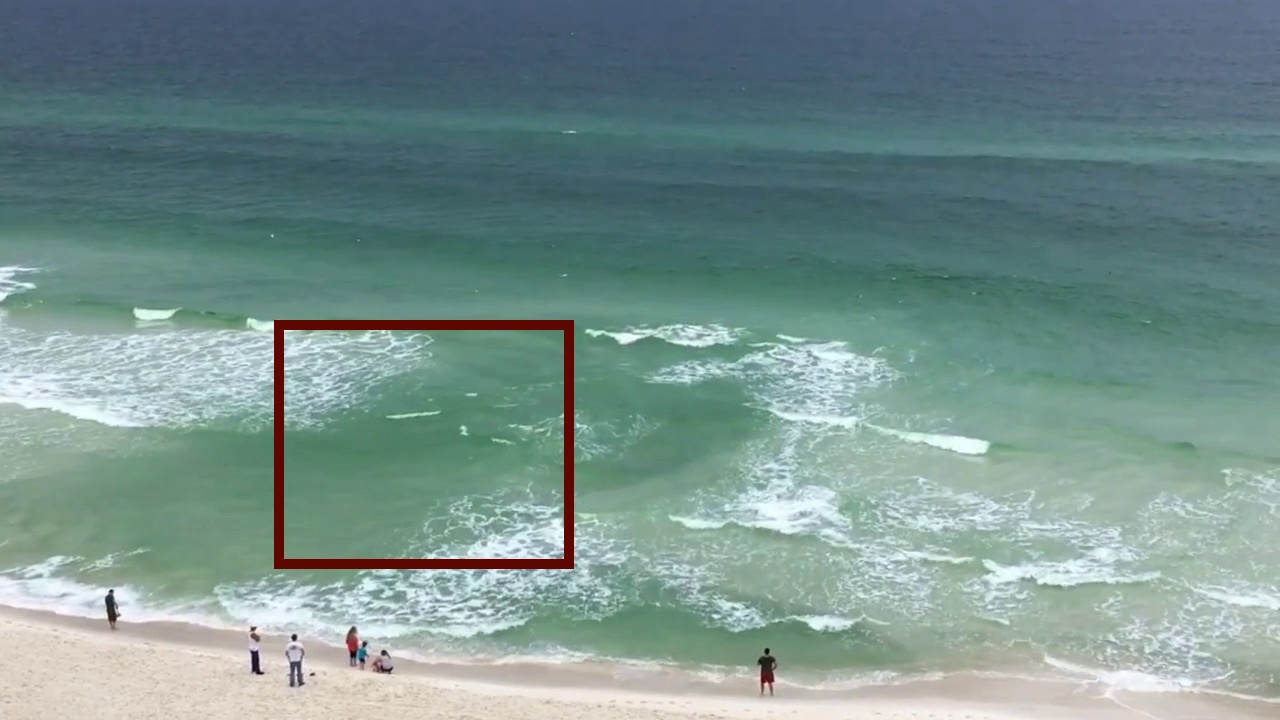}
        \includegraphics[width=0.32\textwidth, height =1.1in]{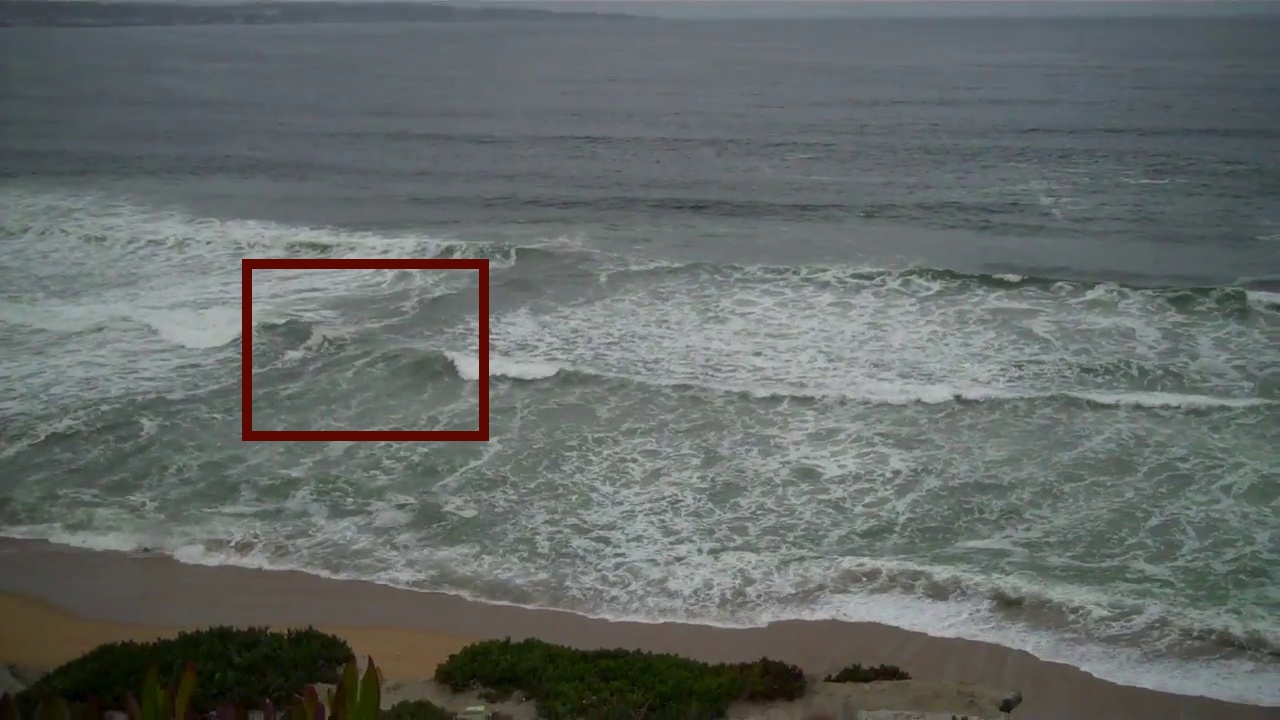}
        \includegraphics[width=0.32\textwidth, height =1.1in]{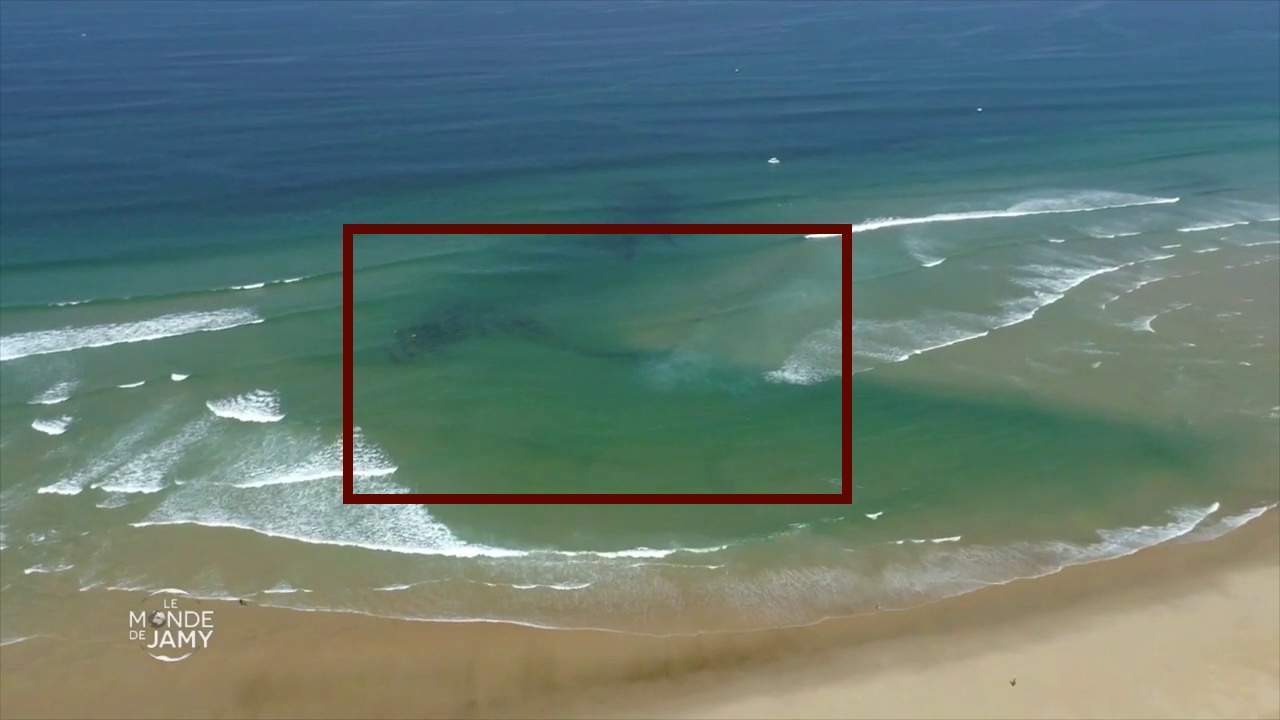}
        \includegraphics[width=0.32\textwidth, height =1.1in]{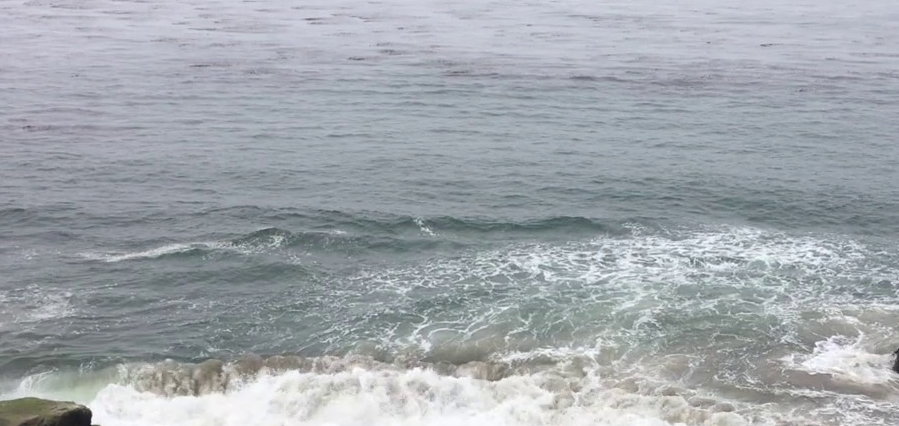}
         \includegraphics[width=0.32\textwidth, height =1.1in]{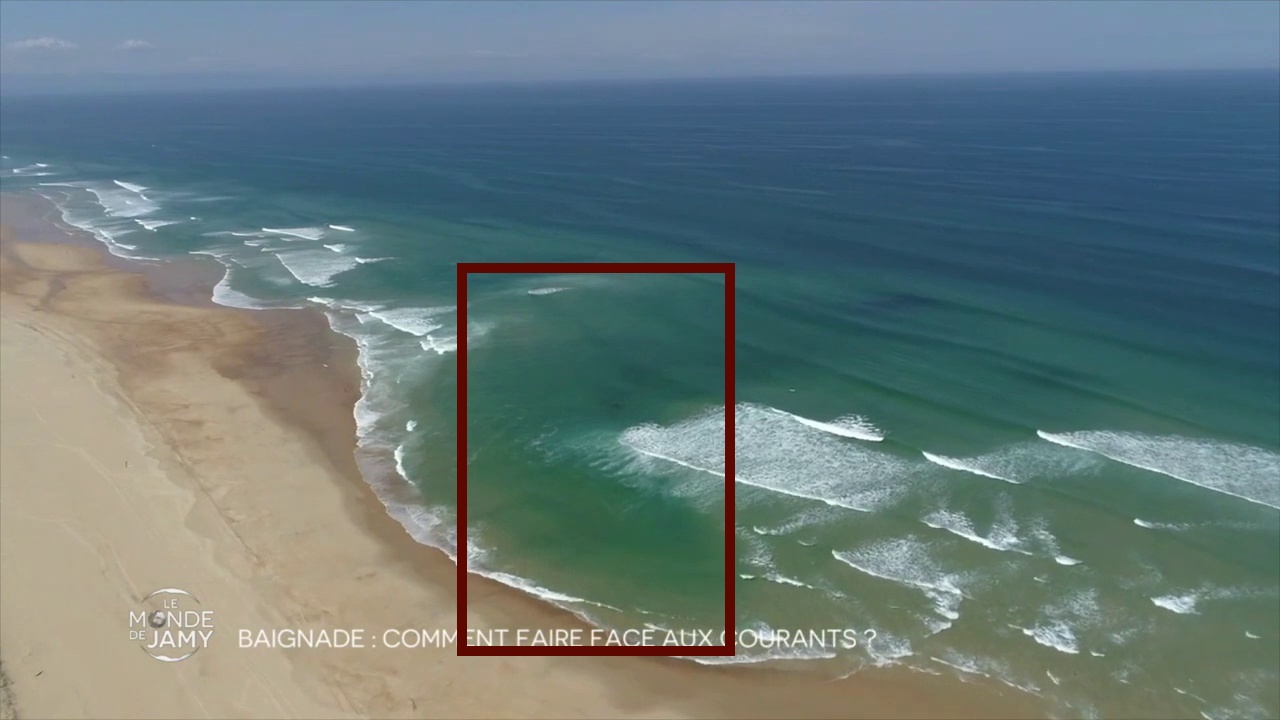}
        \includegraphics[width=0.32\textwidth, height =1.1in]{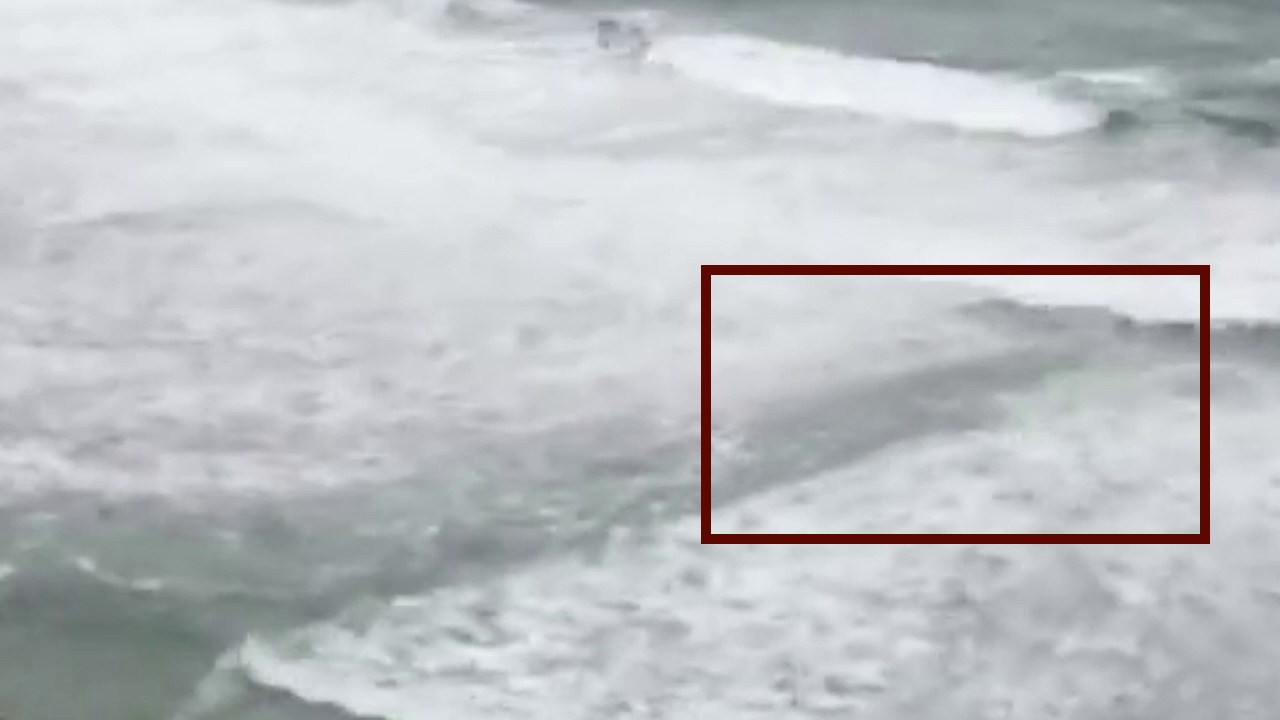}
        \includegraphics[width=0.32\textwidth, height =1.1in]{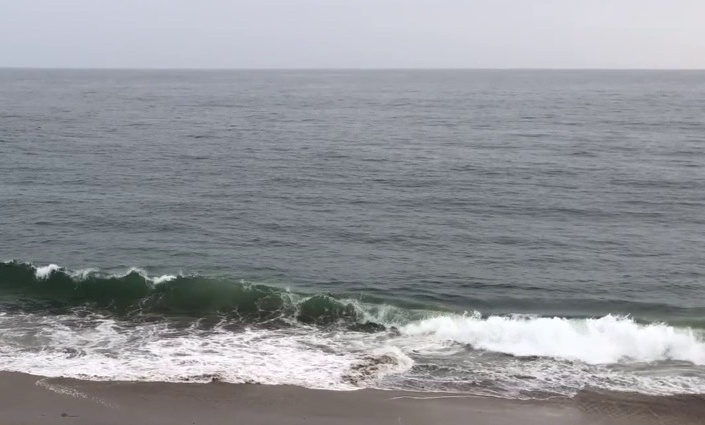}
        \includegraphics[width=0.32\textwidth, height =1.1in]{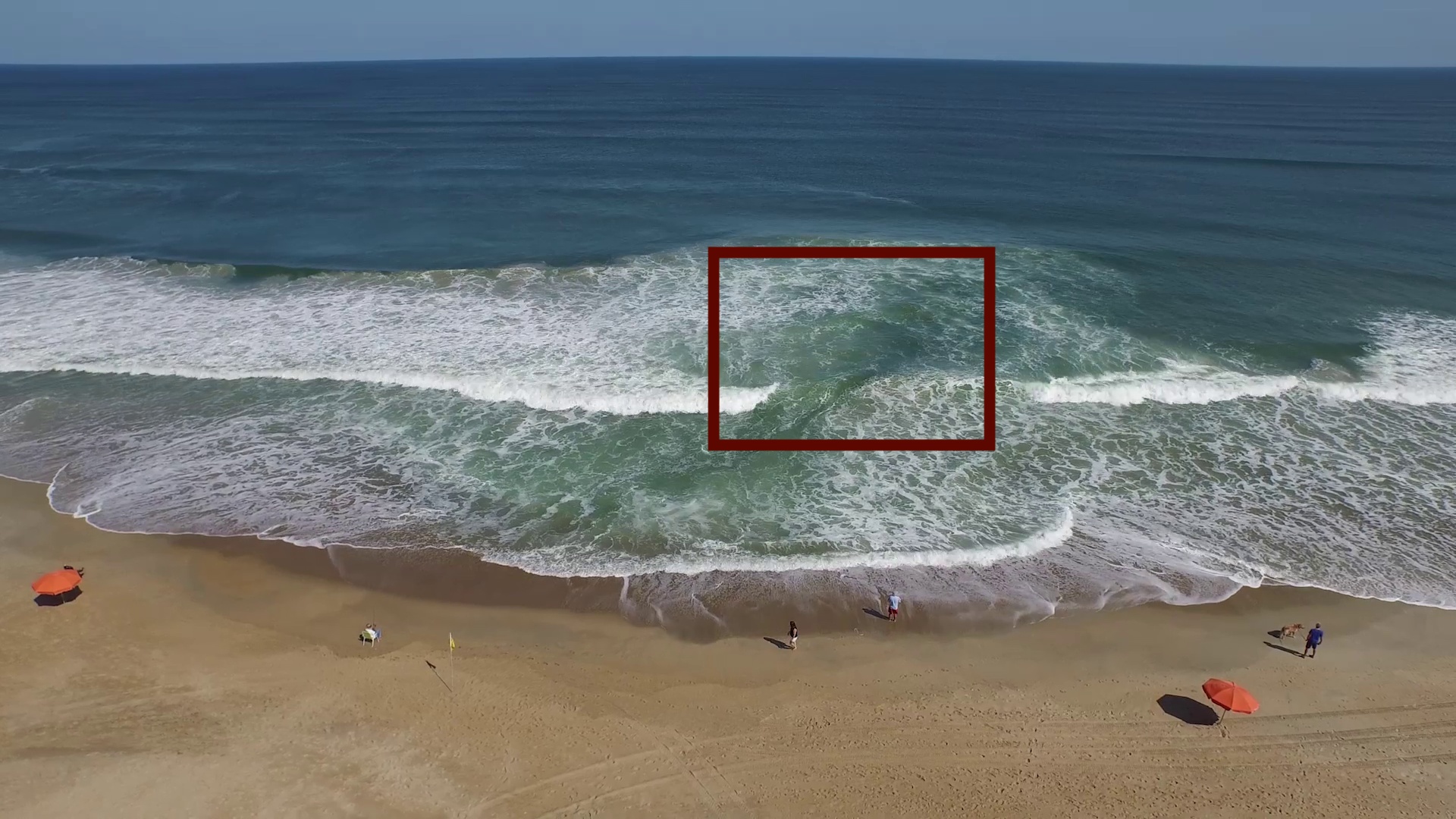}
         \caption{Rip current detections on some of the frames from the test data set. Bounding boxes show the correctly detected rip currents. Frames without bounding boxes do not contain any rip currents.} \label{teaserfig_output}

\end{center}%
\end{figure*}

\newcommand{\colsize}{1.6cm}
\begin{table*}
\small
 \centering

\begin{tabular}{P{3cm} | P{\colsize} P{\colsize} P{\colsize} P{\colsize} P{\colsize} P{\colsize}}. 
    & \text{Human}        & \text{Philip \cite{eurovisshort.20161155}}    & \text{Maryan \cite{maryan2019machine} }     & \text{Maryan}   & \text{F-RCNN}       & \text{F-RCNN+FA}     \\
    & \text{}   & \text{}       & \text{ }  & \text{[modified]}     & \text{[ours]}       & \text{[ours]}     \\
    \hline
\text{$rip\_01.mp4$}  & 0.976 & 0.895  & 0.358 & 0.460 & 0.966 & 1.000 \\
\text{$rip\_02.mp4$}  & 0.700 & 0.098 & 0.698 & 0.135 & 0.776 & 0.860 \\
\text{$rip\_03.mp4$}  & 0.231 & 0.347 & 0.194 & 0.540 & 0.831 & 0.950 \\
\text{$rip\_04.mp4$}  & 0.757 & 0.800 & 0.487 & 0.780 & 0.939 & 0.970 \\
\text{$rip\_05.mp4$}  & 0.883 & 0.280 & 0.736 & 0.450 & 0.834 & 0.957 \\
\text{$rip\_06.mp4$}  & 0.881 & 0.000 & 0.167 & 0.470 & 0.753 & 0.890 \\
\text{$rip\_08.mp4$}  & 0.492 & 0.063 & 0.328 & 0.730 & 0.860 & 0.850 \\
\text{$rip\_11.mp4$}  & 0.824 & 0.000  & 0.563 & 0.940 & 0.930 & 0.951 \\
\text{$rip\_12.mp4$}  & 1.000 & 0.000 & 0.734 & 1.000 & 1.000 & 1.000 \\
\text{$rip\_15.mp4$}  & 0.967 & 0.137 & 0.315 & 0.390 & 0.760 & 0.870\\ 
\text{$rip\_16.mp4$}  & 0.614 & 0.073 & 0.468 & 0.640 & 0.820 & 0.920 \\
\text{$rip\_17.mp4$}  & 1.000 & 0.321 & 0.064 & 0.750 & 0.980 & 1.000 \\
\text{$rip\_18.mp4$}  & 0.563 & 0.218 & 0.250 & 0.240 & 0.790 & 0.890 \\
\text{$rip\_21.mp4$}  & 0.901 & 0.543 & 0.486 & 0.180 & 0.940 & 1.000 \\
\text{$rip\_22.mp4$}  & 0.583 & 0.000 & 0.479 & 0.395 & 0.880 & 0.974 \\
\text{$no\_rip\_01.mp4$}  & 0.986 & 0.169 & 0.000 & 0.972 & 0.813 & 1.000 \\
\text{$no\_rip\_02.mp4$}  & 1.000 & 0.789 & 0.000 & 0.985 & 0.807 & 1.000 \\
\text{$no\_rip\_03.mp4$}  & 0.919 & 0.000 & 0.000 & 0.981 & 0.984 & 1.000 \\
\text{$no\_rip\_04.mp4$}  & 0.952 & 0.000 & 0.000 & 0.974 & 0.835 & 1.000 \\
\text{$no\_rip\_05.mp4$}  & 0.903 & 0.000 & 0.000 & 0.986 & 0.833 & 1.000 \\
\text{$no\_rip\_06.mp4$}  & 1.000 & 0.246 & 0.000 & 0.986 & 0.875 & 1.000 \\
\text{$no\_rip\_07.mp4$}  & 0.983 & 0.525 & 0.000 & 0.982 & 0.875 & 1.000 \\
\text{$no\_rip\_11.mp4$}  & 0.988 & 0.198 & 0.000 & 0.964 & 0.924 & 1.000 \\
\hline
\text{$average\,\, accuracy$} & 0.760 & 0.307 & 0.210 & 0.729 & 0.884& \textbf{0.984} 
\end{tabular}
\caption{Accuracy for each video in the test set. Column 3: Maryan\cite{maryan2019machine} is trained on their training data. Column 4: Maryan [modified] is trained on our training data. Our method has higher overall accuracy than humans or any of the prior methods tested. Frame aggregation does contribute to improvement in accuracy. }
\label{test_data_suite}
\end{table*}



\section{\label{s:results} Results}

We compared the accuracy of our method with human observers as well as two prior methods. We also compared our own method with and without temporal aggregation. 

 \boldheading{Comparison metric. }All methods were tested using the video data set. Frames were labeled as correctly \linebreak classified if the detected bounding boxes have an Intersection over Union (IoU) score versus ground truth above 0.3. IoU is calculated as  $area\_of\_intersection/area\_of\_union$ of the ground truth and the detected bounding boxes. Accuracy of the video was computed as $correct\_labels/total\_frames$, and Table~\ref{test_data_suite} provides the results for all methods. 

\boldheading{Humans.} The primary reason for an automated method of rip current detection is that most people are not good at identifying rip currents~\cite{Brannstrom15}. 
Human annotators were asked to draw bounding boxes around places they believe to have rip currents. We sampled approximately every tenth frame from our video test set and randomized presentation order across all positive and negative examples. Human annotators were not carefully trained, instead they were provided three positive and three negative examples, roughly the amount of information which might fit on a sign at the beach. Annotators were acquired using Mechanical Turk, with basic screening for reliable workers, and paid \$0.10 per image. 

Although human performance was relatively poor with only 76\% of frames labeled correctly, it was higher than we expected based on past studies~\cite{Brannstrom15}. We hypothesize that the clearly labeled sample images on a web page directly adjacent to the image they were asked to label was a positive contributing factor.


\boldheading{Time averaged images.} Maryan et al. \cite{maryan2019machine} perform detection on time averaged images using a boosted cascade of simple Haar like features \cite{Viola2004}. We used Maryan's time averaged data for training since our training data set consists of only static frames.  Testing was performed by first computing time averaged images on each video in our test data set. This method did not perform well. In order to determine if the cause was the images available for training or the model itself, we repeated the experiment with new data. We replaced the relatively small number of low resolution time-averaged images from \cite{maryan2019machine} with the static images from our training data set. Testing this time was against single frames in our test videos. This modification is called Maryan[modified] in the results table. When using our training data the model accuracy improved considerably, leading us to conclude that appropriate training data is critical to good results.  Furthermore, it suggests further investigation on whether using crisp images, rather than time averaged images, to train a model might produce more accurate results.


In our test images/videos the beach is always located at the bottom half of the image/video. However, the training data used by \cite{maryan2019machine} was cropped from images where the beach is located in the top half of the image. Therefore, we were concerned that maybe the difference of orientation between the training data and the test data contributed to the low accuracy of the model. However, when we retrained the model with vertically flipped training data we did not see any significant difference in accuracy on our test images/videos. We hypothesize that the reason for this is that the cropped training images contain insignificant amount of beach pixels.


\boldheading{Optical flow.} Philip et al. \cite{eurovisshort.20161155} compute optical flow on video sequences and make the simplifying assumption that
rip currents can be identified by regions with the second most predominant flow direction, after that
of the primary incoming wave direction, and that they flow in a single seaward direction.
This results in regions of actual rip currents, but also picks up swash regions
where water is washed up the beach and back out to sea with the passing of each wave.
This method was introduced with the primary intention of providing visualizations to users, rather than automated detection. To allow comparison, we modify the method to return a bounding box around the largest detected region, ignoring smaller regions which are less likely to be correct. This method performed poorly on our test videos. We noticed that in videos where there is not enough textural information on the rip current, the optical flow field generated was weak, leading to either missed detections or detection in other regions of the video with stronger texture. 



\boldheading{Frame aggregation.} We implemented frame aggregation as a post process to Faster RCNN initially to temporally stabilize detections, driven by a need for user interpretable visualization of rip current location.  

In order to understand whether temporal smoothing also increased accuracy we analyzed our implementation both with and without frame aggregation. We found that temporal aggregation leads to higher accuracy than using Faster RCNN alone. Example detection results are shown in Figure \ref{teaserfig_output}. Numerical comparison of humans,  prior methods, and our model are provided in Table~\ref{test_data_suite}. Faster RCNN with frame aggregation had the highest accuracy in nearly all cases, and the highest overall (last column of Table~\ref{test_data_suite}). For visual comparisons we have added all the results in the supplementary materials at appendix \ref{s:appendix}.





\section{\label{s:limitations_and_discussion} Discussion and Future Work}

\begin{figure}
\begin{center}
   \includegraphics[width=0.49\linewidth, height=0.95in]{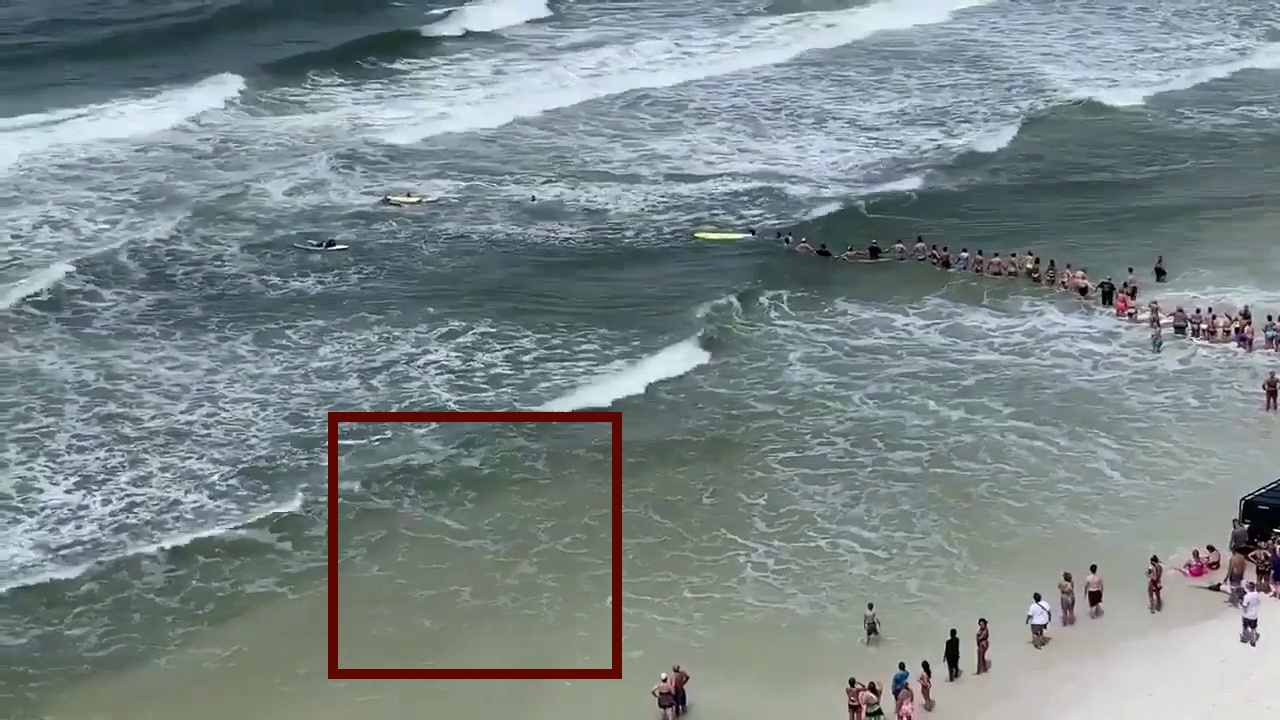}
    \includegraphics[width=0.49\linewidth, height=0.95in]{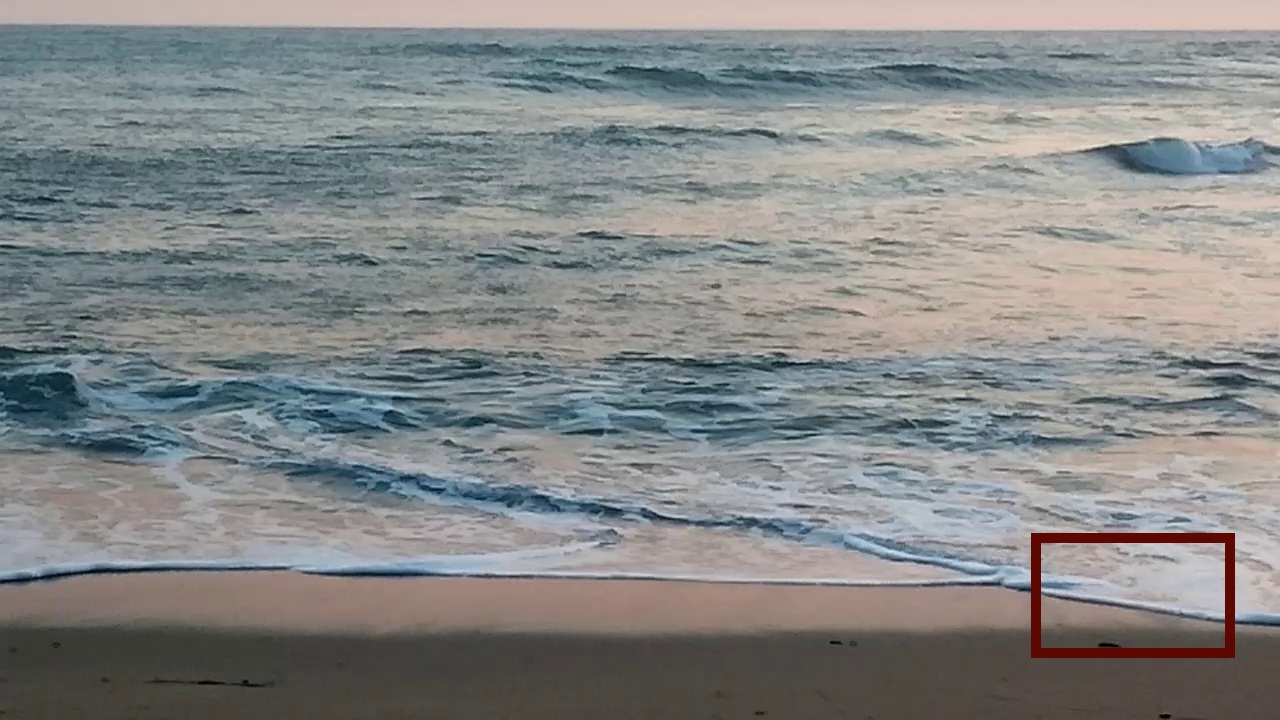}
\end{center}
  \caption{Example failure cases. The false positives on the beach scene (right) are not easily explainable. The false positive on the left scene happens only on spurious frames, which is then corrected by frame aggregation. }
\label{fig:unexpected}
\end{figure}

As with all machine learning models, our implementation can fail when used with images that do not resemble the training data set. Our data sets included primarily rip currents characterized by a gap in  breaking waves, the most common visual indicator for bathymetry controlled rip currents. Thus we would expect to miss rip currents with other visual indicators like sediment plumes. We also expect to fail when presented with new imagery, and occasionally for no apparent reason at all, as seen in Figure \ref{fig:unexpected}.

We found it  difficult to compare our method with prior work and verify that our model performs well in all conditions previously researched, due to a lack of public data sets on which to verify our results. In order to ensure that future work has a baseline from which to compare, our data sets with thousands of labeled frames will be made public. Nevertheless our data sets are still limited. The accuracy numbers presented in this paper are correct on this limited data, but almost certainly overstate probable outcomes in real world deployment. We expect that future work will need to collect more examples including less common rip current visual presentation, a greater variety of scales, and a wider array of beach distractors.

Lastly, our work lacks a success metric which is meaningful to real users. Certainly IoU, true positive rate, mAP, and the like are common in computer vision research, but is it appropriate to measure accuracy on single frames? Most conceivable deployed systems, such as pointing a mobile phone at the beach, have access to video. If measuring accuracy on video, should we measure accuracy aggregated over 1 second or 1 minute? Is a tight bounding box measured by IoU needed, or just a general region? A metric useful to researchers and simultaneously meaningful to users is needed.

\section{\label{s:conclusion}Conclusion}     

We present a machine learning approach for identifying rip currents automatically. We use Faster RCNN and a custom temporal aggregation stage to make detections from still images or videos with higher measured accuracy than both humans and other methods of rip current detection previously reported in the literature.
Training data set and suite of test videos are made available for other researchers.

\appendix 
\section{\label{s:appendix} Appendix} 
Supplementary material to this article can be found online at \url{https://sites.google.com/view/ripcurrentdetection/home}



\bibliographystyle{cas-model2-names}

\bibliography{main}

\end{document}